\documentclass{article}
\usepackage[utf8]{inputenc}
\usepackage{setspace}
\usepackage[margin=1.1in]{geometry}
\usepackage{graphicx}
\graphicspath{ {./figures/} }

\usepackage[utf8]{inputenc} 
\usepackage[T1]{fontenc}    
\usepackage{hyperref}       
\usepackage{url}            
\usepackage{booktabs}       
\usepackage{amsfonts}       
\usepackage{nicefrac}       
\usepackage{microtype}      
\usepackage{xcolor}         
\usepackage{bm}
\usepackage{graphicx}
\usepackage{parskip}
\usepackage{algorithmic}

\usepackage{amsthm}

\newtheorem{sublemma}{Lemma}
\usepackage{algorithm}
\usepackage{mathtools}
\usepackage{amssymb}
\usepackage{wrapfig}
\usepackage{setspace}
\usepackage{xcolor} 

\usepackage{placeins}
\usepackage{tikz}
\usetikzlibrary{fit,calc}

\usepackage{bbm}

\usepackage{amsmath,amsfonts,bm}

\def\eqref#1{equation~\ref{#1}}

\def\1{\bm{1}}

\def\rmP{{\mathbf{P}}}

\DeclareMathAlphabet{\mathsfit}{\encodingdefault}{\sfdefault}{m}{sl}
\SetMathAlphabet{\mathsfit}{bold}{\encodingdefault}{\sfdefault}{bx}{n}

\newcommand{\E}{\mathbb{E}}

\usepackage{xpatch}
\makeatletter
\AtBeginDocument{\xpatchcmd{\@thm}{\thm@headpunct{.}}{\thm@headpunct{}}{}{}}
\makeatother

\title{Minimax Demographic Group Fairness in Federated Learning}

\author{%
  Afroditi Papadaki \\
  {\normalsize University College London}\\
  \and
   Natalia Martinez\\
  {\normalsize Duke University} \\
   \and
   Martin Bertran\\
  {\normalsize Duke University} \\
   \and
   Guillermo Sapiro \\
   {\normalsize Duke University and Apple Inc.} \\
  \and
   Miguel Rodrigues\\
  {\normalsize University College London} \\
}

\date{\vspace{-6ex}}

\begin{document}

\maketitle

{\begin{center}
    \texttt{\{a.papadaki.17, m.rodrigues\}@ucl.ac.uk}\\
\texttt{\{natalia.martinez, martin.bertran, guillermo.sapiro\}@duke.edu}\\

\end{center}}

\begin{abstract}
Federated learning is an increasingly popular paradigm that enables a large number of entities to collaboratively learn better models. In this work, we study minimax group fairness in {federated learning} {scenarios} where different participating entities may only have access to a subset of the population groups during the training phase. We formally analyze how {our proposed} group fairness objective differs from existing federated learning fairness criteria that impose similar performance across participants instead of demographic groups. We provide an optimization algorithm -- FedMinMax -- for solving the proposed problem that provably enjoys the performance guarantees of centralized learning algorithms. We experimentally compare the proposed approach against other {state-of-the-art} methods in terms of group fairness in various federated learning setups{, showing that our approach exhibits competitive or superior performance.}
\end{abstract}


\section{Introduction}
Machine learning models are being increasingly adopted to make decisions in a range of domains, such as finance, insurance, medical diagnosis, recruitment, and many more \cite{10.1145/3376898}. Therefore, we are often confronted with the need -- sometimes imposed by regulatory bodies -- to ensure that such machine learning models do not lead to decisions that discriminate individuals from a certain demographic group.

\begin{wrapfigure}{r}{0.65\textwidth} 
    \centering
    \includegraphics[width=8.5cm]{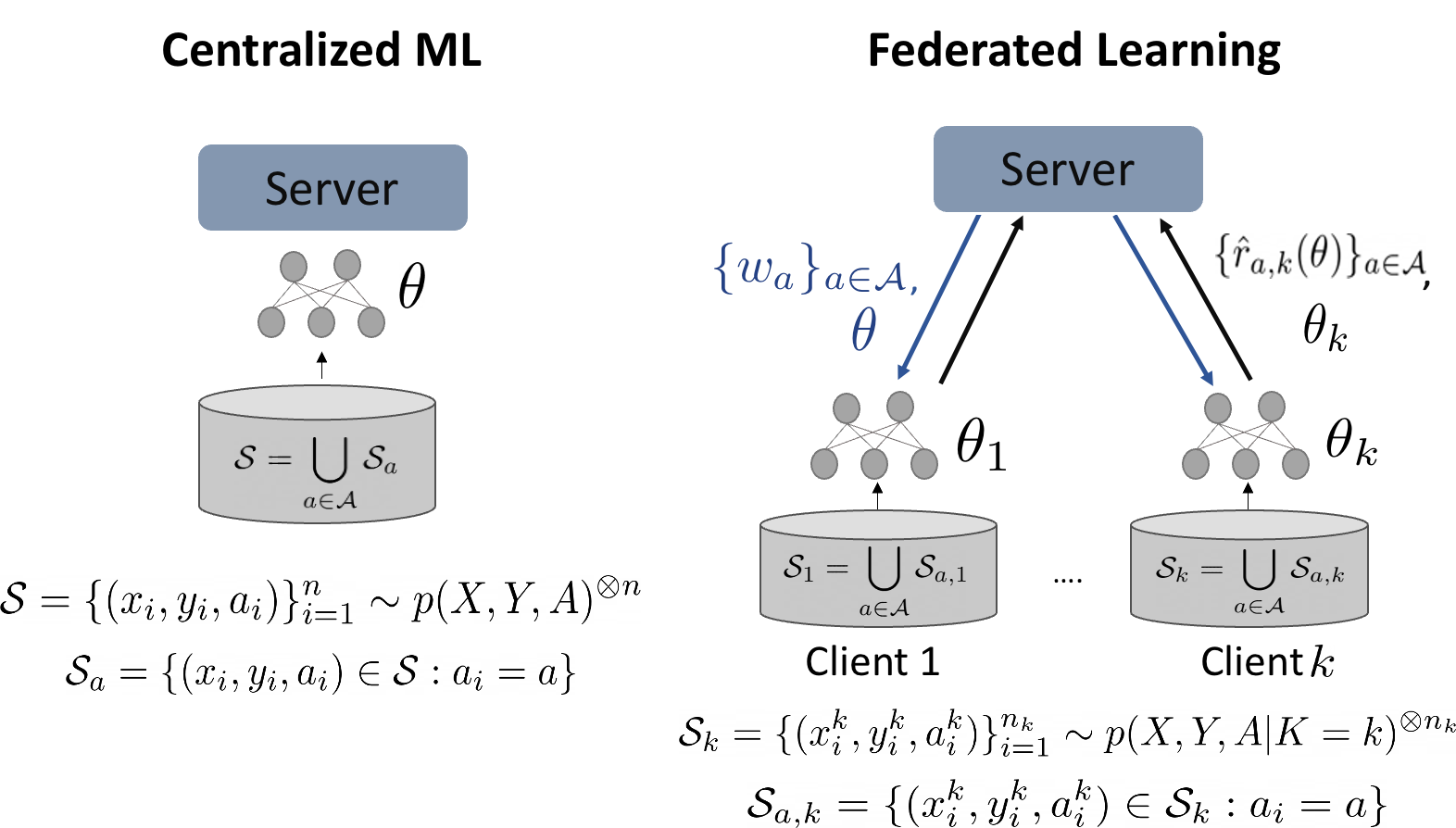}
    
    \caption{Centralized Learning vs. Federated Learning group fairness. \textit{Left:} A single entity holds the dataset $\mathcal{S}$ in a single server that is responsible for learning a model $h$ parameterized by $\theta$.  \textit{Right:} Multiple entities hold different datasets $\mathcal{S}_k$, sharing restricted information with a server that is responsible for learning a model $h$ parametrized by $\bm \theta$, {and the group importance weights $\bm w=\{w_a\}_{a \in \mathcal{A}}$.} See also Section 3.}
    
    \label{fig:Fairschemes}
\end{wrapfigure}

The development of machine learning models that are fair across different (demographic) groups has been well studied in traditional learning setups where there is a single entity responsible for learning a model based on a local dataset holding data from individuals of the various groups. 
However, there are settings where the data representing different demographic groups is spread across multiple entities rather than concentrated on a single entity/server. For example, consider a scenario where various hospitals wish to learn a diagnostic machine learning model that is fair (or performs reasonably well) across different demographic groups but each hospital may only contain training data from certain groups because -- in view of its geo-location -- it serves predominantly individuals of a given demographic \cite{cui2021addressing}. 
This new setup along with the conventional centralized one are depicted in Figure \ref{fig:Fairschemes}.

These emerging scenarios however bring about various challenges. The first challenge relates to the fact that each individual entity may not be able to learn locally by itself a fair machine learning model because it may not hold (or hold little) data from certain demographic groups{. The second challenge} relates to that fact that each individual entity may also not be able to directly share their own data with other entities due to legal or regulatory challenges such as GDPR \cite{EUdataregulations2018}. Therefore, the conventional machine learning fairness \textit{ansatz} -- relying on the fact that the learner has access to the overall data --  does not generalize from the centralized data setup to the new distributed one.

It is possible to address these challenges by adopting federated learning (FL) approaches. These learning approaches enable multiple entities (or clients\footnote{Clients are different user devices, organisations or even geo-distributed datacenters of a single company \cite{advancesFL}. In this manuscript we use the terms participants, clients, and entities, interchangeably.}) coordinated by a central server to iteratively learn in a decentralized manner a single global model to carry out some task~\cite{DBLP:journals/corr/KonecnyMRR16,DBLP:journals/corr/KonecnyMYRSB16}. The clients do not share data with one another or with the server; instead the clients only share focused updates with the server, the server then updates a global model, and distributes the updated model to the clients, with the process carried out over multiple rounds or iterations. This learning approach enables different clients with limited local training data to learn better machine learning models.

However, with the exception of some recent works such as \cite{cui2021addressing,zhang2021unified}, which we will discuss later, federated learning is not typically used to learn models that exhibit performance guarantees for different demographic groups served by a client (i.e., \textit{group fairness} guarantees); instead, it is primarily used to learn models that exhibit specific performance guarantees for each client involved in the federation (i.e., \textit{client fairness} guarantees). Importantly, in view of the fact that a machine learning model that is \textit{client fair} is not necessarily \textit{group fair} (as we formally demonstrate in this work), it becomes crucial to understand how to develop new federated learning techniques leading up to models that are also fair across different demographic groups.

This work develops a new federated learning algorithm that can be adopted by multiple entities coordinated by a single server to learn a \textit{global minimax group fair} model. We show that our algorithm leads to the same (minimax) group fairness performance guarantees of centralized approaches such as \cite{diana2020convergent,martinez2020minimax}, which are exclusively applicable to settings where the data is concentrated in a single client. Interestingly, this also applies to scenarios where certain clients do not hold any data from some of the demographic groups.

The rest of the paper is organized as follows: Section \ref{sec:related_work} overviews related work. Section \ref{sec:prob_form} formulates our proposed distributed group fairness problem. Section \ref{clientvsgroupfairness} formally demonstrates that traditional federated learning approaches such as \cite{DBLP:journals/corr/abs-2108-08435,DRFA,Li2020Fair,AFL} may not always solve group fairness. In Section \ref{sec:algo} we propose a new federated learning algorithm 
to collaboratively learn models that are minimax group fair. Section \ref{sec:experiemtns} illustrates the performance of our approach in relation to other baselines.  
Finally, Section \ref{sec:conclusion} draws various conclusions.

\section{Related Work}\label{sec:related_work}
\noindent\textit{\textbf{Fairness in Machine Learning.} }The development of fair machine learning models in the standard \textit{centralized learning setting} -- where the learner has access to all the data -- is underpinned by fairness criteria. One popular criterion is \textit{individual fairness} \cite{dwork2011fairness} that dictates that the model is fair provided that people with similar characteristics/attributes are subject to similar model predictions/decisions.
Another family of criteria -- known as \textit{group fairness} -- requires the model to perform similarly on different demographic groups. Popular  group fairness criteria include equality of odds, equality of opportunity \cite{hardt2016equality}, and demographic parity \cite{louizos2017variational}, that are usually imposed as a constraint within the learning problem.
More recently, \cite{martinez2020minimax} introduced \textit{minimax group fairness}; this criterion requires the model to optimize the prediction performance of the worst demographic group without unnecessarily impairing the performance of other demographic groups (also known as no-harm fairness) \cite{diana2020convergent,martinez2020minimax}. 
In this work we leverage minimax group fairness criterion to learn a model that is (demographic) group fair across any groups included in the clients distribution in federated learning settings. {However,}  {the overall concepts here introduced can also be extended to other fairness criteria.}

\smallskip
\noindent\textit{\textbf{Fairness in Federated Learning.}} The development of fair machine learning models in \textit{federated learning settings} has been building upon the group fairness literature. 
The majority of these works has concentrated predominantly on \textit{client-fairness} which targets the development of algorithms leading to models that exhibit similar performance across different clients~\cite{Li2020Fair}.

One such approach is agnostic federated learning (AFL) \cite{AFL}, whose aim is to learn a model that optimizes the performance of the worst performing client. Extensions of AFL \cite{DRFA,afedavg} improve its communication-efficiency by enabling clients to perform multiple local optimization steps.
Another FL approach proposed in \cite{Li2020Fair}, uses an extra fairness constraint to flexibly control performance disparities across clients. Similarly, tilted empirical risk minimization \cite{li2021tilted} uses a hyperparameter called tilt to enable fairness or robustness by magnifying or suppressing the impact of individual client losses. FedMGDA$+$ \cite{fedmgda+} is an algorithm that combines minimax optimization coupled with Pareto efficiency \cite{micro_theory} and gradient normalization to ensure fairness across users and robustness against malicious clients. 

The works in \cite{horvath2021fjord,munir2021fedprune} enable fairness across clients with different hardware computational capabilities by allowing any participant to train a submodel of the original {deep neural network (DNN)} in order to contribute to the global model.
The authors in \cite{Wang2021FederatedLW} observe that unfairness across clients is caused by conflicting gradients that may significantly reduce the performance of some clients and {therefore} propose an algorithm for detecting and mitigating such conflicts.
Finally, GIFAIR-FL \cite{DBLP:journals/corr/abs-2108-0274} uses a regularization term to penalize the spread in the aggregated loss to enforce uniform performance across the participating entities.
Our work naturally departs from these fairness federated learning approaches since, as we prove in Section \ref{clientvsgroupfairness}, client-fairness ensures fairness across all demographic groups included across clients datasets {only} under some special conditions.

Another fairness concept in federated learning is collaborative fairness \cite{fan2021improving,9378043,Lyu2020,Nagalapatti_Narayanam_2021}, which proposes each client's performance compensation to correspond to its contribution on the utility task of the global model. Larger rewards to high-contributing clients motivate their participation in the federation while lower rewards prevent free-riders \cite{Lyu2020}.
However, such approaches {might} further penalize clients that have access to the worst performing demographic groups resulting to a even more unfair global model. 

There are some recent complementary works that consider group fairness within client distributions. Group distributional robust optimization (G-DRFA) \cite{zhang2021unified}, aims to optimize for the worst performing group by learning a weighting coefficient for each local group, even if there are shared groups across clients.
In our work, we combine the statistics received from the clients sharing the same groups to learn a global model, since, as we experimentally show in Section \ref{sec:experiemtns}, considering duplicates of the same group might lead to worst generalization in some FL scenarios.
FCFL \cite{cui2021addressing} focuses on improving the worst performing client while ensuring a level of local group fairness defined by each client, by employing gradient-based constrained multi-objective optimization. 
Our primary goal is to learn a model solving (demographic) group fairness across any groups included in the clients distribution, independently of the groups representation in a particular client. 

{Finally,} some recent approaches study the effects of (demographic) group fairness in FL using metrics such as demographic parity and/or equality in opportunity \cite{DBLP:journals/corr/abs-2010-05057,cui2021addressing,rodriguezgalvez2021enforcing,zeng2021improving,ezzeldin2021fairfed,DBLP:journals/corr/abs-2105-11570,chu2021fedfair}. Compared to these methods, our approach can support scenarios with multiple group attributes and targets without any modifications on the optimization procedure. Also, even though comparing different fairness metrics is out of the scope of this work,\footnote{There are various studies discussing the effects of different fairness metrics. See for example \cite{10.1145/3287560.3287589}.} the aforementioned methods enforce some type of zero risk disparity across groups\footnote{The risk considered in the fairness constraints is different across fairness definitions.} and thus degrade the performance of the good performing groups. In this work, we consider minimax group fairness criterion \cite{martinez2020minimax,diana2020convergent}, and due to its no-unnecessary harm property, we do not disadvantage any demographic groups except if absolutely necessary, making it suitable for applications such as healthcare and finance. 
{Our formulation is complemented by theoretical results connecting minimax client and minimax group fairness and by proposing a {provably convergent} optimization algorithm.}

\smallskip

\noindent\textit{\textbf{Robustness in Federated Learning.}} Works dealing with robustness to distributional shifts in user data, such as \cite{scaffold,NEURIPS2020_f5e53608}, also relate to group fairness. One work that closely relates to group fairness is FedRobust \cite{NEURIPS2020_f5e53608}, {that aims to learn a model for the worst case affine shift, by assuming that a client's data distribution is an affine transformation of a global one.} However, it requires each client to have enough data to estimate the local worst case shift else the global model performance on the worst group hinders \cite{li2021fedbn}.
\smallskip

\noindent
{{\bf Our Contributions.} To recap, our core contributions compared to the literature are:

\begin{itemize}

\item We formulate minimax group fairness in federated learning settings where some clients might only have access to a subset of the demographic groups during the training phase.

\item We formally show under what conditions minimax group fairness is equivalent to minimax client fairness so that optimizing for any of the two notions results into a model that is both group and client fair. 

\item We propose a provably convergent optimization algorithm to collaboratively learn a minimax fair model across any demographic groups included in the federation, that allows clients to have high, low or no representation of a particular group. We show that our federated learning algorithm leads to a global model that is equivalent to a model yielded by a centralized learning algorithm. 
\end{itemize}
}

\section{Problem Formulation}\label{sec:prob_form}
\subsection{Group Fairness in Centralized Machine Learning}

We first describe the standard minimax group fairness problem in a centralized machine learning setting \cite{diana2020convergent,martinez2020minimax}, where there is a single entity/server holding all relevant data and responsible for learning a group fair model (see Figure \ref{fig:Fairschemes}). We concentrate on classification tasks, though our approach also applies to other learning tasks such as regression. 
 Let the triplet of random variables $(X,Y,A) \in \mathcal{X} \times \mathcal{Y} \times \mathcal{A}$ represent input features, target, and demographic groups. Let also $p(X,Y,A)=p(A) \cdot p(X,Y|A)$ represent the joint distribution of these random variables where $p(A)$ represents the prior distribution of the different demographic groups and $p(X,Y|A)$ their data conditional distribution.

Let $\ell:\Delta^{|\mathcal{Y}|-1} \times \Delta^{|\mathcal{Y}|-1} \to {\rm I\!R}_+$ be a loss function where $\Delta$ represents the probability simplex.
We now consider that the entity will learn an hypothesis $h$ drawn from an hypothesis class $\mathcal{H}=\{h:\mathcal{X} \rightarrow \Delta^{|\mathcal{Y}|-1} \}$, that solves the optimization problem given by
\begin{equation}\label{objective_centralized}
    \min_{h \in \mathcal{H}} \max_{a \in \mathcal{A}} r_{a}(h)\textnormal{,  } r_a(h)= \mathop{\E}_{(X,Y)\sim p(X,Y|A=a)}[\ell(h(X),Y)|A=a].
\end{equation}
Note that this problem involves the minimization of the expected risk of the worst performing demographic group.

Importantly, under the assumption that the loss is a convex function w.r.t the hypothesis\footnote{This is true for the most common functions in machine learning settings such as Brier score and cross entropy.} and the hypothesis class is a convex set, solving the minimax objective in Eq. \ref{objective_centralized} is equivalent to solving

\begin{equation}\label{objective_centralized_linear}
    \min_{h \in \mathcal{H}} \max_{a \in \mathcal{A}} r_{a}(h) \ge  \min_{h \in \mathcal{H}} \max_{\mu \in \Delta^{|\mathcal{A}|-1}_{\ge \epsilon}} \sum_{a\in\mathcal{A}}\mu_a r_a(h) 
\end{equation}
where $\Delta_{\ge \epsilon}^{|\mathcal{A}| - 1}$ represent the vectors in the simplex with all of their components larger than $\epsilon$. 
Note that if $\epsilon=0$ the inequality in Eq. \ref{objective_centralized_linear} becomes an equality, however, allowing zero value coefficients may lead to models that are weakly, but not strictly, Pareto optimal \cite{geoffrion1968proper,miettinen2012nonlinear}. 

The minimax objective over the {linear combination of the sensitive groups} can be achieved by alternating between projected gradient ascent or multiplicative weight updates to optimize the weights {given the model,} and stochastic gradient descent to optimize the model {given the weighting coefficients} \cite{chen2018,diana2020convergent,martinez2020minimax}.

\subsection{Group Fairness in Federated Learning}

We now describe our proposed group fairness federated learning problem; this problem differs from the previous one because the data is now distributed across multiple clients but each client (or the server) do not have direct access to the data held by other clients. See also Figure \ref{fig:Fairschemes}.

In this setting, we incorporate {a} categorical variable $K \in \mathcal{K}$ {to} our data tuple $(X,Y,A,K)$ to indicate the clients participating in the federation. The joint distribution of these variables is $p(X,Y,A,K)=p(K) \cdot p(A|K) \cdot p(X,Y|A,K)$, where $p(K)$ represents a prior distribution over clients -- which in practice is the fraction of samples that are acquired by client $K$ relative to the total number of data samples --, $p(A|K)$ {represents the distribution of the groups conditioned on the client, and $p(X,Y|A,K)$ represents the distribution of the distribution of the input and target variables conditioned on the group and client.} 
We assume that the group-conditional distribution is the same across clients, meaning $p(X,Y|A,K) = p(X,Y|A)$. Note, however, that our model explicitly allows for the distribution of the demographic groups to depend on the client (via $p(A|K)$), accommodating for the fact that certain clients may have a higher (or lower) representation of certain demographic groups over others.

We now aim to learn a model $h \in \mathcal{H} $ that solves the minimax fairness problem as presented in Eq. \ref{objective_centralized}, but considering that the group loss estimates are split into $|\mathcal{K}|$ estimators associated {with} each client. We therefore re-express the linear weighted formulation of Eq. \ref{objective_centralized_linear} using importance weights, allowing to incorporate the role of the different clients, as follows:
\begin{equation}\label{fed_objective_convex}
\begin{array}{ll}
     \min\limits_{h \in \mathcal{H}} \max\limits_{\mu \in \Delta^{|\mathcal{A}|-1}_{\ge \epsilon}} \sum\limits_{a\in\mathcal{A}}\mu_ar_a(h) =& 
    
    \min\limits_{h \in \mathcal{H}} \max\limits_{\mu \in \Delta^{|\mathcal{A}|-1}_{\ge \epsilon}}\sum\limits_{a \in \mathcal{A}} p(A=a) w_a r_a(h) 
     \\
     &= \min\limits_{h \in \mathcal{H}} \max\limits_{\mu \in \Delta^{|\mathcal{A}|-1}_{\ge \epsilon}}\sum\limits_{k \in \mathcal{K}}  p(K=k) \sum\limits_{a \in \mathcal{A}} p(A=a|K=k) w_a r_a(h)
         \\
     &= \min\limits_{h \in \mathcal{H}} \max\limits_{\mu \in \Delta^{|\mathcal{A}|-1}_{\ge \epsilon}}\sum\limits_{k \in \mathcal{K}}  p(K=k) r_k(h,{\bm w}),
\end{array}
\end{equation}
where $r_k(h,{\bm w})=\sum\limits_{a \in \mathcal{A}} p(A=a|K=k) w_a r_a(h)$ is the expected client risk and $w_a=\mu_a/p(A=a)$ denotes the importance weight for a particular {demographic} group.

There is an immediate non-trivial challenge that arises within this proposed federated learning setting in relation to the centralized one described earlier: we need to devise an algorithm that solves the objective in Eq. \ref{fed_objective_convex} under the constraint that the different clients cannot share their local data with the server or with one another, but -- in line with conventional federated learning settings \cite{DRFA,Li2020Fair,DBLP:journals/corr/McMahanMRA16,AFL}-- only {local model updates} of a global model (or other quantities such as local risks) are shared with the server. This will be addressed later in this paper by the proposed federated optimization.

\section{Client Fairness vs. Group Fairness in Federated Learning}\label{clientvsgroupfairness}

Before proposing a federated learning algorithm to solve our proposed group fairness problem, we first reflect whether a model that solves the more widely used client fairness objective in federated learning settings given by \cite{AFL}
\begin{equation}
    \min\limits_{h \in \mathcal{H}}\max\limits_{k \in \mathcal{K}}r_{k}(h) = \min\limits_{h \in \mathcal{H}}\max\limits_{\bm\lambda \in \Delta^{|\mathcal{K}|-1}} \mathop{\E}_{\mathcal{D}_{\bm\lambda}} [\ell(h(X),Y)], \label{client_fairness_objective} 
\end{equation}
where $\mathcal{D}_{\bm\lambda}=\sum\limits_{k=1}^{|\mathcal{K}|} \lambda_k p(X,Y|K=k)$ denotes a joint data distribution over the clients and {$\bm \lambda=\{ \lambda_k\}_{k \in \mathcal{K}}$ is the vector consisting of client weighting coefficients}, also solves our proposed minimax group fairness objective given by
\begin{equation}
     \min\limits_{h \in \mathcal{H}}\max\limits_{a \in \mathcal{A}}r_{a}(h) = \min\limits_{h \in \mathcal{H}}\max\limits_{\bm\mu \in \Delta^{|\mathcal{A}|-1}} \mathop{\E}_{\mathcal{D}_{\bm\mu}}[\ell(h(X),Y)], \label{group_fairness_objective} 
\end{equation}
where $\mathcal{D}_{\bm\mu}=\sum\limits_{a=1}^{|\mathcal{A}|} \mu_a p(X,Y|A=a)$ denotes a joint data distribution over sensitive groups and {$\bm \mu=\{ \mu_a\}_{a \in \mathcal{A}}$ is the vector of the group weights}.

The following lemma illustrates that a model that is minimax fair with respect to the clients is equivalent to a relaxed minimax fair model with respect to the (demographic) groups.

\begin{sublemma}\label{lemma1}

Let $\rmP_{\mathcal{A}}$ denote a matrix whose entry in row $a$ and column $k$ is $p(A=a|K=k)$ (i.e., the prior of group $a$ in client $k$). 
Then, given a solution to the minimax problem across clients
\begin{equation}
    h^*,\bm\lambda^* \in \arg\min_{h \in \mathcal{H}}\max_{\bm\lambda \in \Delta^{|\mathcal{K}|-1}} \mathop{\E}_{\mathcal{D}_{\bm \lambda}}[\ell(h(X),Y)], 
\end{equation}
$\exists$ $\bm\mu^*=\rmP_{\mathcal{A}}\bm\lambda^*$ that is solution to the following constrained minimax problem across sensitive groups:
\begin{equation}
     h^*,\bm\mu^* \in \arg\min_{h \in \mathcal{H}}\max_{\bm \mu \in \rmP_{\mathcal{A}}\Delta^{|\mathcal{K}|-1}}\mathop{\E}_{\mathcal{D}_{\bm \mu}}[\ell(h(X),Y)],
\end{equation}

\noindent where the weighting vector $\bm\mu$ is constrained to belong to the simplex subset defined by $\rmP_{\mathcal{A}}\Delta^{|\mathcal{K}|-1} \subseteq \Delta^{|\mathcal{A}|-1}$.
In particular, if the set ${\Gamma} = \big\{ \bm\mu' \in \rmP_{\mathcal{A}}\Delta^{|\mathcal{K}|-1}$: ${\bm\mu'} \in \arg\min\limits_{h \in \mathcal{H}}\max\limits_{ {\bm \mu} \in \Delta^{|\mathcal{A}|-1}}\mathop{\E}_{\mathcal{D}_{\bm \mu}}[\ell(h(X),Y)]\big\} \not= \emptyset$, then ${\bm \mu}^* \in \Gamma $, and the minimax fairness solution across clients is also a minimax fairness solution across demographic groups. 
\end{sublemma}

Lemma \ref{lemma1} proves that being minimax with respect to the clients is equivalent to finding the group minimax model constraining the weighting vectors $\bm\mu$ to be inside the simplex subset $\rmP_{\mathcal{A}}\Delta^{|\mathcal{K}|-1}$. Therefore,
if this set already contains a group minimax weighting vector, then the group minimax model is equivalent to client minimax model.
Another way to interpret this result is that being minimax with respect to the clients is the same as being minimax for any group assignment $\mathcal{A}$ such that linear combinations of the groups distributions are able to generate all clients distributions, and there is a group minimax weighting vector in $\rmP_{\mathcal{A}}\Delta^{|\mathcal{N}|-1}$. 

Being minimax at the client and group level relies on $\rmP_{\mathcal{A}}\Delta^{|\mathcal{K}|-1}$ containing the minimax weighting vector. In particular, if for each sensitive group there is a client comprised entirely of this group ($\rmP_{\mathcal{A}}$ contains a identity block), then $\rmP_{\mathcal{A}}\Delta^{|\mathcal{K}|-1} = \Delta^{|\mathcal{A}|-1}$ and group and client level fairness are guaranteed to be fully compatible. Another trivial example is when at least one of the client's group priors is equal to a group minimax weighting vector. This result also suggests that client level fairness may also differ from group level fairness. This motivates us to develop a new federated learning algorithm to guarantee group fairness that -- where the conditions of the lemma hold -- also results in client fairness.  We experimentally validate the insights deriving from Lemma \ref{lemma1} in Section \ref{sec:experiemtns}. The proof for Lemma \ref{lemma1} {is provided in the }supplementary material, Appendix \ref{lemma_1_proof}.

\section{MiniMax Group Fairness Federating Learning Algorithm}\label{sec:algo}

We now propose an optimization algorithm -- Federated Minimax (FedMinMax) -- to solve the group fairness problem in Eq. \ref{fed_objective_convex}. 

We let each client $k$ have access to a dataset $\mathcal{S}_k = \{(x_i^{k},y_i^{k},a_i^{k}); i=1,\ldots,n_k\}$ containing various data points drawn i.i.d according to $p(X,Y,A|K=k)$. We also define three additional sets: (a) $\mathcal{S}_{a,k}=\{(x_i^{k},y_i^{k},a_i^{k}) \in \mathcal{S}_k: a_i=a\}$ is a set containing all data examples associated with group $a$ in client $k$; (b) $\mathcal{S}_{a}=\bigcup\limits_{k \in \mathcal{K}} \mathcal{S}_{k,a}$ is the set containing all data examples associated with group $a$ across the various clients; and (c) {$\mathcal{S} = \bigcup\limits_{k \in \mathcal{K}} \mathcal{S}_{k} = \bigcup\limits_{a \in \mathcal{A}} \mathcal{S}_{a} = \bigcup\limits_{k \in \mathcal{K}} \bigcup\limits_{a \in \mathcal{A}} \mathcal{S}_{a,k}$} is containing all data examples across groups and across clients. 
Note again that -- in view of our modelling assumptions -- it is possible that $\mathcal{S}_{a,k}$ can be empty for some $k$ and some $a$ implying that such a client does not have data realizations for such group.

We will also let the model $h$ be parameterized via a vector of parameters $ \bm \theta \in \Theta$, i.e., $h(\cdot) = h(\cdot;\bm \theta)$. \footnote{This vector of parameters could for example correspond to the set of weights / biases in a neural network.} Then, one can approximate the relevant statistical risks using empirical risks as 

\begin{equation}\label{empirical_group_risk}
\begin{array}{cc}
     \hat{r}_k(\bm \theta, { \bm w}) = \sum\limits_{a \in \mathcal{A}}\frac{n_{a,k}}{n_k} {\hat w}_a \hat{r}_{a,k}(\bm \theta),& \hat{r}_a (\bm \theta) = \sum\limits_{k \in \mathcal{K}} \frac{n_{a,k}}{n_a} \hat{r}_{a,k}(\bm \theta),  
\end{array}
\end{equation}
where $\hat{r}_{a,k} (\bm \theta) = \frac{1}{n_{a,k}}\sum\limits_{(x,y) \in \mathcal{S}_{a,k}} \ell (h(x;\bm \theta), y)$,  ${\hat w}_a=\mu_a /(n_a/n)$, $n_k = |\mathcal{S}_k|$, $n_a = |\mathcal{S}_a|$, $n_{a,k} = |\mathcal{S}_{a,k}|$, and $n = |\mathcal{S}|$. Note that $\hat{r}_k(\bm \theta, \bm w)$ is an estimate of $r_k(\bm \theta, \bm w)$, $\hat{r}_a(\bm \theta)$ is an estimate of {$r_a(\bm \theta)$}, and $\hat{r}_{a,k} (\bm \theta)$ is an estimate of {$r_{a,k}(\bm \theta)=\mathop{\E}_{(X,Y)\sim p(X,Y|A=a,K=k)}[\ell(h(X),Y)|A=a,K=k]$.}

We consider the importance weighted empirical risk ${\hat r}_k$ since the clients do not have access to the data distribution but instead to a dataset with finite samples. 
Therefore, the clients in coordination with the central server attempt to solve the optimization problem given by:
\begin{equation}\label{empirical_fed_objective}
\min_{\bm \theta \in \Theta} \max_{\bm\mu \in \Delta_{\ge \epsilon}^{|
\mathcal{A}| - 1}} \hat{r}_a(\bm \theta)  \coloneqq \sum_{a \in \mathcal{A}} \mu_a \hat{r}_a(\bm \theta)
\; \; \; \equiv \; \; \; 
\min_{\bm \theta \in \Theta} \max_{\bm\mu \in \Delta_{\ge \epsilon}^{|
\mathcal{A}| - 1}}  \sum_{k \in \mathcal{K}} \frac{n_k}{n} \hat{r}_k(\bm \theta, { \bm w}).
\end{equation}

The objective in Eq. \ref{empirical_fed_objective} can be interpreted as a zero-sum game between two players: the learner aims to minimize the objective by optimizing the model parameters $\bm \theta$ and the adversary seeks to maximize the objective by optimizing the weighting coefficients $\bm \mu$.

We use a non-stochastic variant of the stochastic-AFL algorithm introduced in \cite{AFL}. 
Our version, provided in Algorithm \ref{alg:fedminmax}, assumes that all clients are available to participate in each communication round $t$. In each round $t$, the clients receive the latest model parameters $\bm \theta^{t-1}$, the clients then perform one gradient descent step using all their available data, and the clients then share the updated model parameters along with certain empirical risks with the server. The server (learner) then performs a weighted average of the client model parameters $\bm \theta^{t} = \sum\limits_{k \in \mathcal{K}} \frac{n_k}{n} \bm \theta^{t}_k$. 
The server also updates the weighting coefficient using a projected gradient ascent step in order to guarantee that the weighting coefficient updates are consistent with the constraints. We use the Euclidean algorithm proposed in \cite{10.1145/1390156.1390191} in order to implement the projection operation ($\prod_{\Delta^{|\mathcal{A}|-1}} (\cdot)$).


\begin{algorithm}[ht]
\footnotesize
    \caption{\footnotesize\textsc{Federated MiniMax (FedMinMax) }}
    \label{alg:fedminmax}
     {\bfseries Input:} $\mathcal{K}$: Set of clients, $T:$ total number of communication rounds,
     $\eta_{\bm \theta}$: model learning rate, $\eta_{\bm\mu}$: global adversary learning rate, $\mathcal{S}_{a,k}$: set of examples for group $a$ in client $k$, $\forall a \in \mathcal{A}$ and $\forall k \in \mathcal{K}$.

    \begin{algorithmic}[1]
    \setstretch{1.35}
        \STATE Server {\bfseries initializes} {$\bm\mu^0\leftarrow \bm \rho = \{|\mathcal{S}_a|/|\mathcal{S}| \}_{a \in \mathcal{A}}$} and $\bm \theta^0$ randomly.
    
    \FOR{$t=1$ {\bfseries to} $T$}

    \STATE Server {\bfseries computes} {$\bm w^{t-1} \leftarrow \bm \mu^{t-1} /\bm \rho$}
    
    \STATE Server {\bfseries broadcasts} $\bm \theta^{t-1}$, $\bm w^{t-1}$ 

      \FOR{each client $k \in \mathcal{K}$ {\bfseries in parallel }}
        \STATE  $\bm \theta^{t}_k \leftarrow \bm \theta^{t-1} - \eta_\theta \nabla_\theta \hat{r}_k(\bm \theta^{t-1}, \bm w^{t-1})$%
     
        \STATE Client-$k$ {\bfseries obtains} and {\bfseries sends} $\{\hat{r}_{a,k}(\bm \theta^{t-1})\}_{a \in \mathcal{A}}$ and  $\bm \theta^{t}_k$ to server
        \ENDFOR 
        
    \STATE Server {\bfseries computes}:  $\bm \theta^{t} \leftarrow \sum\limits_{k \in \mathcal{K}} \frac{n_k}{n} \bm \theta^{t}_k$
     
     \STATE Server {\bfseries updates:}
     $\bm\mu^{t} \leftarrow \prod_{\Delta^{|\mathcal{A}|-1}} \Big (\bm{{\mu}}^{t-1} +\eta_{\bm{\mu}} {\nabla}_{\bm{\mu}}\langle\ \bm \mu^{t-1},\hat{r}_a(\bm \theta^{t-1})\rangle \Big ) $ 
     \ENDFOR
    
    \end{algorithmic}
    {\bfseries Outputs}: $\frac{1}{T} \sum_{t =1}^T \bm \theta^t$
    \end{algorithm}    
    

We can show that our proposed algorithm can exhibit convergence guarantees.

\begin{sublemma}\label{centralized_vs_federated_solution}
Consider our federated learning setting (Figure \ref{fig:Fairschemes}, right) where each entity $k$ has access to a local dataset $\mathcal{S}_k = \bigcup\limits_{a \in \mathcal{A}} \mathcal{S}_{a,k}$, and a centralized machine learning setting (Figure \ref{fig:Fairschemes}, left) where there is a single entity that has access to a single dataset $\mathcal{S} = \bigcup\limits_{k \in \mathcal{K}} \mathcal{S}_k = \bigcup\limits_{k \in \mathcal{K}} \bigcup\limits_{a \in \mathcal{A}} \mathcal{S}_{a,k}$ (i.e., this single entity in the centralized setting has access to the data of the various clients in the distributed setting). 
Then, Algorithm \ref{alg:fedminmax} (federated) and Algorithm \ref{alg:centralized_minmax} (non-federated, in supplementary material, Appendix \ref{centralized_algo}) lead to the same global model provided that learning rates and model initialization are identical.
\end{sublemma}

The proof for Lemma \ref{centralized_vs_federated_solution} is provided in Appendix \ref{lemma_2_proof}. This lemma shows that our federated learning algorithm inherits any convergence guarantees of existing centralized machine learning algorithms. In particular, assuming that one can model the single gradient descent step using a $\delta$-approximate Bayesian Oracle~\cite{chen2018}, we can show that a centralized algorithm converges and hence our FedMinMax one converges too (under mild conditions on the loss function, hypothesis class, and learning rates). See Theorem 7 in ~\cite{chen2018}.

\section{Experimental Results}\label{sec:experiemtns}
\begin{figure}[H]
\centering
\includegraphics[width=13cm]{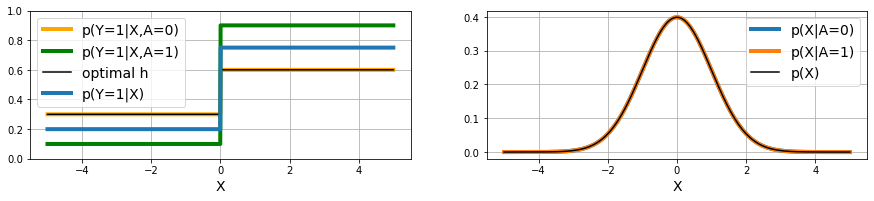}
\setlength{\abovecaptionskip}{-5pt}

\caption{Illustration of the optimal hypothesis $h$ and the conditional distributions $p(Y|X)$ and $p(X|A)$ for the generated synthetic dataset.{ (\textit{Left}:) The worst group is $A=0$ and the minimax optimal hypothesis $h$ (black line) is equal to the optimal model for the worst group (orange line). (\textit{Right}:) The distributions $p(X)$, and conditional distributions $p(X|A=0)$ and $p(X|A=1)$ are overlapping.}}

\label{fig:synthetic_data}
\end{figure}

In this section we empirically showcase the applicability and competitive performance of the proposed federated learning algorithm.
We apply FedMinMax to diverse federated learning scenarios {by utilizing} common benchmark datasets with multiple targets and sensitive groups. In particular, we perform experiments on the following datasets:
\begin{itemize}
    \item \textbf{Synthetic.} We generated a synthetic dataset for binary classification involving two sensitive groups (i.e., $|\mathcal{A}|=2$). {Let $\mathcal{N}(\mu,\sigma^2)$ be the normal distribution with $\mu$ being the mean and $\sigma^2$ being the variance, and $Ber(p)$ Bernoulli distribution with probability $p$.} The data were generated assuming the group variable $A \sim Ber(\frac{1}{2})$, the input features variable $X \sim \mathcal{N}(0,1)$ and the target variable $Y|X,A=a\sim Ber(h^*_a)$, where $h^*_a=u^l_a \mathbbm{1}[x \leq0]+u_a^h \mathbbm{1}[x >0]$ is the optimal hypothesis for group $A=a$. We select $\{u_0^h,u_1^h,u_0^l,u_1^l\}=\{0.6,0.9,0.3,0.1\}$. As illustrated in Figure \ref{fig:synthetic_data}, left side, the optimal hypothesis $h$ is equal to the optimal model for group $A=0$. 
    
    \item \textbf{Adult \cite{adult}.} Adult is a binary classification dataset consisting of $32,561$ entries for predicting yearly income based on twelve input features such as age, race, education and marital status. We consider four sensitive groups (i.e., $|\mathcal{A}|=4$) created by combining the gender labels and the yearly income as follows: \{Male w/ income $>50$K, Male w/ income $<=50$K, Female w/ income $>50$K, Female w/ income $<=50$K\}.
    
    \item \textbf{FashionMNIST \cite{fashionmnist}.} FashionMNIST is a grayscale image dataset which includes $60,000$ training images and $10,000$ testing images. The images consist of $28\times28$ pixels and are classified into 10 clothing categories. In our experiments we consider each of the target categories to be a sensitive group too, (i.e., $|\mathcal{A}|=10$). 
    
    \item \textbf{CIFAR-10 \cite{cifar}.} CIFAR-10 is a collection of $60,000$ colour images of  $32\times32$ pixels. Each image contains one out of 10 object classes. There are $50,000$ training images and $10,000$ test images. We use all ten target categories, which we assign both as targets and sensitive groups (i.e., $|\mathcal{A}|=10$). 
    
    \item {\textbf{ACS Employment \cite{ding2021retiring}.} ACS Employment is a recent dataset constructed using ACS PUMS data for predicting whether an individual is employed or not. For our experiments we use the 2018 1-Year data for all the US states and Puerto Rico. We combine race and utility labels to generate the following sensitive groups: : \{Employed White, Employed Black, Employed Other, Unemployed White, Unemployed Black, Unemployed Other\} (i.e., $|\mathcal{A}|=6$). We also conduct experiments where the sensitive class is race using the original 9 labels that we report in supplementary material, Appendix \ref{additional_results}.
}
\end{itemize}

\smallskip

We also examine three federated learning {settings}, that we categorize based on the sensitive group allocation on clients as follows:
\begin{enumerate}
    \item \textit{Equal access to Sensitive Groups (ESG)}, where every client has access to all sensitive groups but does not have enough data to train a model individually. Each client in the federation has access to the same amount of the sensitive classes (i.e., $n_i= n_j \forall i,j \in \mathcal{K},i \neq j$ and $n_{a,i}= n_{a,j} \forall i,j \in \mathcal{K}, a \in \mathcal{A}, i \neq j$). Here we examine a case where group and client fairness are not equivalent.
    \item \textit{Partial access to Sensitive Groups (PSG)}, where each participant has access to a subset of the available groups memberships. In particular, the data distribution is unbalanced across participants since the size of local datasets differs (i.e., $n_i\neq n_j \forall i,j \in \mathcal{K},i \neq j$). Akin to ESG, this is a scenario where group and client fairness are incompatible. We use this scenario to compare the performances when there is low or no local representation of particular groups.
    \item  \textit{{Access} to a Single Sensitive Group (SSG)}, {where} each client holds data from one sensitive group, for showcasing the group and client fairness objectives equivalence derived from Lemma \ref{lemma1}. Similarly to PSG setting, the size of the local dataset varies across clients.
\end{enumerate}

Note that ESG is an i.i.d. data scenario while PSG and SSG are non-i.i.d. data settings. 
Also note that each client's data is unique, meaning that there are no duplicated examples across clients.
In all experiments we consider a federation consisting of 40 clients and a single server that orchestrates the training procedure. We benchmark our approach against AFL \cite{AFL}, $q$-FedAvg \cite{Li2020Fair}, TERM \cite{li2021tilted} and FedAvg \cite{DBLP:journals/corr/McMahanMRA16}. Further, as a baseline, we also run FedMinMax with one client (akin to centralized ML), that we denote \textit{Centralized Minmax Baseline}, to confirm Lemma \ref{centralized_vs_federated_solution}. {We do not compare to }baselines that explicitly employ a different fairness metric (e.g., demographic parity) since this not the focus of this work. For all the datasets, we compute the means and standard deviations of the accuracies and risks over three runs. 
We assume that every client is available to participate at each communication round for every method to make the comparison more fair. More details about model architectures and experiments are provided in Appendix \ref{more_results}.

\begin{figure}[ht]
    \centering
    \includegraphics[width=\textwidth]{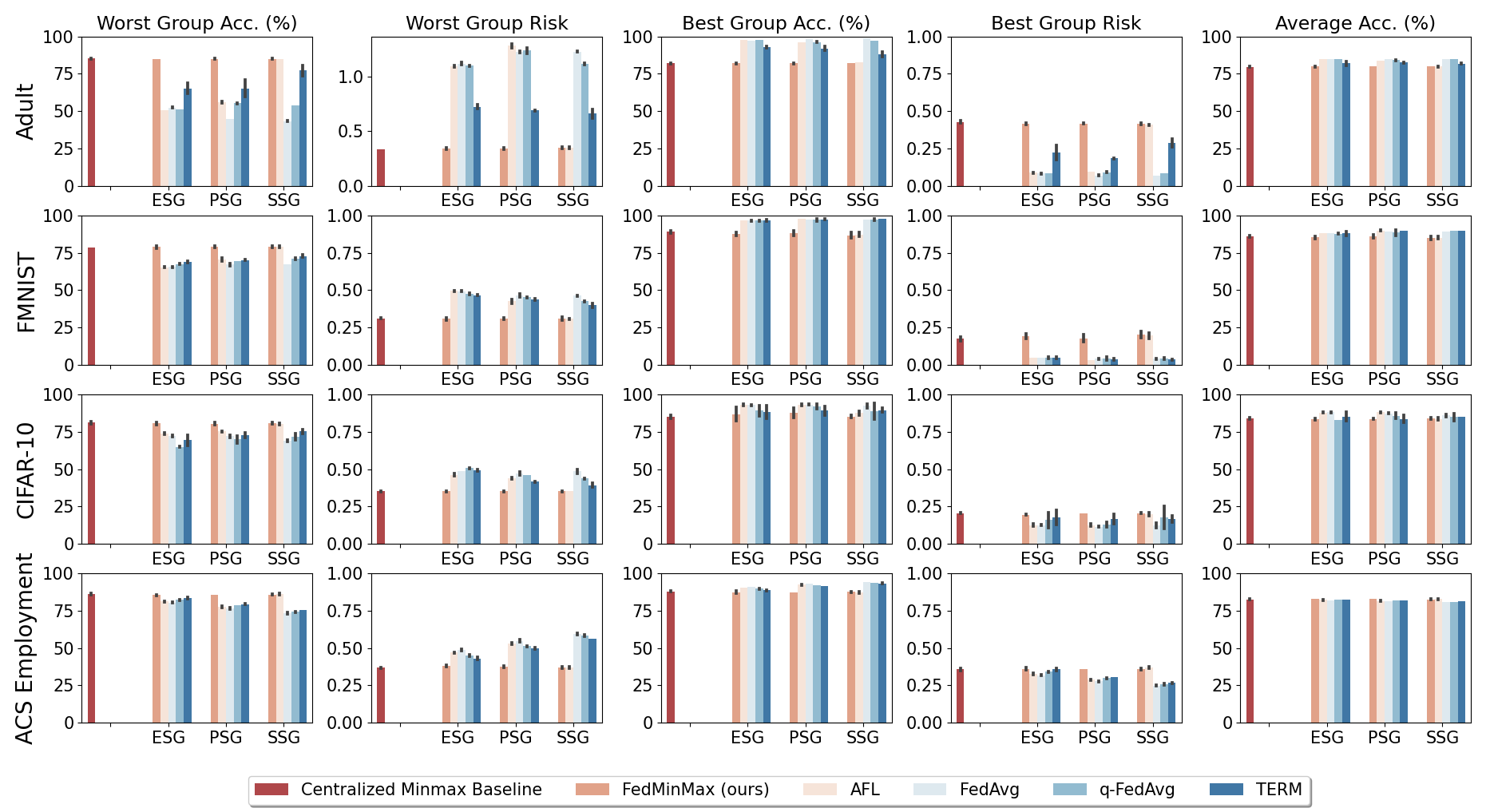}
    \setlength{\abovecaptionskip}{-1pt}
    \caption{Comparison of the worst group, best group, and average risks and accuracies across three runs for AFL, FedAvg, $q$-FedAvg, TERM, FedMinmax and Centralized Minmax Baseline, across the different federated learning scenarios. Each bar reports the mean and standard deviation of the respective metric on the testing set. The numerical values, showing the advantages of the proposed framework, are provided in Tables \ref{tab:synthetic}, \ref{tab:analytic_adult_risk}, \ref{tab:analytic_fmnist}, \ref{analytic_cifar10} and \ref{tab:analytic_acs_race_target_risks}, in the supplementary material. }
    
    \label{fig:accs_comparison}
\end{figure}

We begin by investigating the worst group, the best group and the average utility performance for the Adult, FashionMNIST, CIFAR-10 and ACS Employment datasets in Figure \ref{fig:accs_comparison}. We present the mean and standard deviation of the accuracies and risks on the test dataset.
FedMinMax enjoys a similar accuracy to the Centralized Minimax Baseline in all settings, as proved in Lemma \ref{centralized_vs_federated_solution}. AFL is similar to FedMinMax and Centralized Minmax Baseline only in SSG, where group fairness is implied by client fairness, in line with Lemma \ref{lemma1}. FedAvg has similar best accuracy across federated settings, however the accuracy of the worst group decreases as the local data becomes more heterogeneous (i.e., in PSG and SSG). 
In many datasets, $q$-FedAvg and TERM have superior performance on the worst group compared to AFL and FedAvg in PSG and ESG, but do not to achieve minimax group fairness on any of the FL settings.
Note that FedMinMax has the best worst group performance in all settings as expected. 

For the numerical values, illustrating the efficiency of the proposed approach for every setting and dataset, see Tables \ref{tab:synthetic}, \ref{tab:analytic_adult_risk}, \ref{tab:analytic_fmnist}, \ref{analytic_cifar10} and \ref{tab:analytic_acs_race_target_risks}, in the supplementary material.

\begin{figure}[ht]
    \centering
    \includegraphics[width=\textwidth]{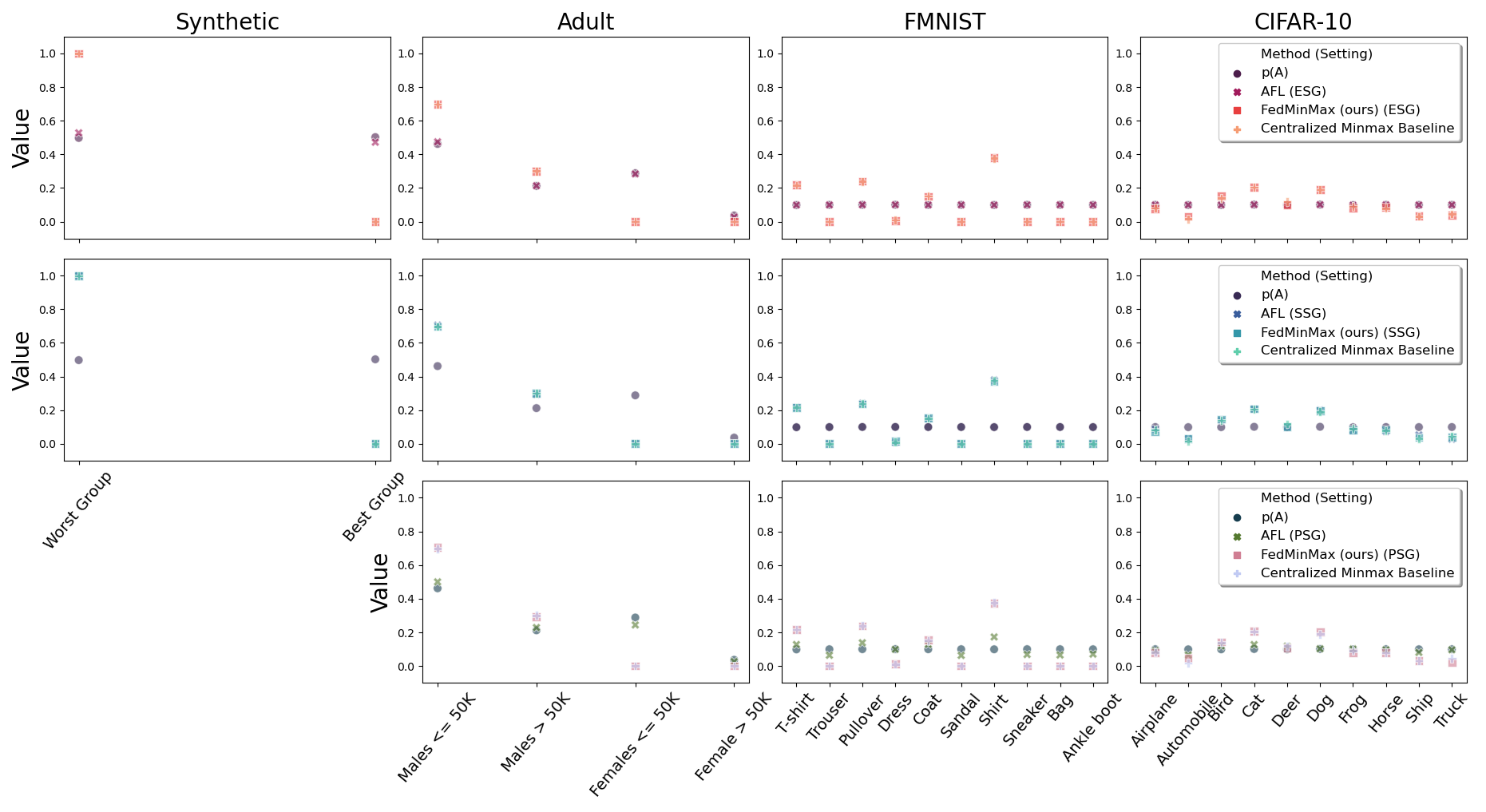}
    \setlength{\abovecaptionskip}{-15pt}
    \caption{Sensitive group weighting coefficients for every minimax approach considered across different datasets calculated during the training time. We also provide the prior group distribution $p(A)$. Note that the weighting coefficients were produced based on the group risks on the training dataset and might not necessarily correspond to the group risks on the test set.}
    \label{fig:coeff_comparison}
\end{figure}

Next we show the final group weighting coefficients for the minimax approaches AFL, FedMinMax, and Centralized Minmax Baseline in Figure \ref{fig:coeff_comparison}. Note that PSG scenario is valid only for datasets where $|\mathcal{A}|>2$, else its equivalent to SSG setting. 

The proposed approach yields similar group weights across all settings. FedMinMax also achieves the same weighting coefficients to Centralized Minmax Baseline, akin to Lemma \ref{centralized_vs_federated_solution}. AFL produces weights similar to the group priors in ESG that move towards the minimax weighting coefficients the more we increase the heterogeneity w.r.t. the sensitive groups.
AFL achieves the similar weights to FedMinMax and Centralized Minmax Baseline only in SSG scenario where each participant has access to exactly one group, following Lemma \ref{lemma1}.
Note that the group weighting coefficients are updated based on the risks calculated on the training set and might not generalize to the testing set for every dataset.
We provide a complete description of the weighting coefficients for each approach in Tables \ref{tab:coeff_synthetic}, \ref{tab:coeff_adult}, \ref{tab:coeff_fmnist}, \ref{tab:coeff_cifar} and \ref{tab:coeff_acs_employment_race_target}, in {the} supplementary material.

 Finally, we illustrate the efficiency of considering global demographics across entities instead of multiple local ones, as in \cite{zhang2021unified}. For these experiments, we re-purpose our algorithm -- we call the adjusted version LocalFedMinMax -- so that the adversary proposes a weighting coefficient for each group located in a client (i.e., $\bm \mu=\{ \{ \mu_{a,k}\}_{a \in \mathcal{A}}\}_{k \in \mathcal{K}}$). Recall that the adversary in our proposed algorithm uses a single weighting coefficient for every common demographic group (i.e., $\bm \mu=\{ \mu_a\}_{a \in \mathcal{A}}$).
We {provide the detailed} description of LocalFedMinMax in Algorithm \ref{alg:localfedminmax}, Appendix \ref{compl_algorithms}.

In Table \ref{tab:summary_local_global} we report results for both approaches on two federations consisting of 10 and 40 participants, respectively.
LocalFedMinMax and FedMinMax offer similar improvement on the worst group on SSG regardless the number of clients. We also notice a similar behavior in the smaller federated network for the ESG scenario. In the remaining settings, LocalFedMinMax, leads to a worst performance as the amount of client increases and {the number of data for each group per client reduces.} On the other hand, FedMinMax is not effected by the local group representation since it aggregates the statistics received by each client and updates the weights for (global) demographics, {leading up to a better generalization performance.}

\begin{table}[ht]
    \centering
        \caption{Comparison of the worst group risk achieved for FedMinMax and LocalFedMinMax on FashionMNIST and CIFAR-10 datasets. We highlight the worst values. Extended versions for both datasets can be found in Tables \ref{tab:local_global_fmnist} and \ref{tab:local_global_cifar10}.}
    \label{tab:summary_local_global}
\resizebox{0.8\textwidth}{!}{
\begin{tabular}{lcccccc}\toprule
 &\multicolumn{6}{c}{\textbf{FashionMNIST}}
\\\cmidrule(lr){2-7} 
 &\multicolumn{3}{c}{$10$ \textbf{Clients}} & \multicolumn{3}{c}{\textbf{$40$ Clients}}
\\\cmidrule(lr){2-4}\cmidrule(lr){5-7}
     \textbf{Method} & \textbf{ESG}  & \textbf{PSG} & \textbf{SSG}    &\textbf{ESG}  & \textbf{PSG} & \textbf{SSG}\\\midrule

LocalFedMinMax    & 0.316$\pm$0.092 & \textbf{0.331$\pm$0.007} & 0.309$\pm$0.013 & \textbf{0.346$\pm$0.081} & \textbf{0.331$\pm$0.021} & 0.31$\pm$0.005 \\
FedMinMax & 0.31$\pm$0.005  &  0.308$\pm$0.012& 0.308$\pm$0.003  & 0.307$\pm$0.01  &  0.31$\pm$0.008 & 0.309$\pm$0.011\\\midrule
&\multicolumn{6}{c}{\textbf{CIFAR-10}}
\\\cmidrule(lr){2-7} 
& \multicolumn{3}{c}{$10$ \textbf{Clients}} & \multicolumn{3}{c}{\textbf{$40$ Clients}}
\\\cmidrule(lr){2-4}\cmidrule(lr){5-7}
     \textbf{Method} & \textbf{ESG}  & \textbf{PSG} & \textbf{SSG}    &\textbf{ESG}  & \textbf{PSG} & \textbf{SSG}\\\midrule

LocalFedMinMax & 0.358$\pm$0.008 & 0.353$\pm$0.042 & 0.352$\pm$0.0 & \textbf{0.381$\pm$0.004} &\textbf{0.378$\pm$0.005} & 0.352$\pm$0.007 \\
FedMinMax &  0.352$\pm$0.02  & 0.351$\pm$0.005 & 0.351$\pm$0.0  & 0.351$\pm$0.002 & 0.351$\pm$0.009 &0.351$\pm$0.002  \\\bottomrule
\end{tabular}}

\end{table}

\section{Conclusion}
In this work, we formulate (demographic) group fairness in federated learning setups where different participating entities may only have access to a subset of the population groups during the training phase (but not necessarily the testing phase), exhibiting minmax fairness performance guarantees akin to those in centralized machine learning settings.

We formally show how our fairness definition differs from the existing fair federated learning works, offering conditions under which conventional client-level fairness is equivalent to group-level fairness.
We also provide an optimization algorithm, FedMinMax, to solve the minmax group fairness problem in federated setups that exhibits minmax guarantees akin to those of minmax group fair centralized machine learning algorithms.

We empirically confirm that our method outperforms existing federated learning methods in terms of group fairness in various learning settings and validate the conditions under which the competing approaches yield the same solution as our objective.\label{sec:conclusion}

\newpage

\bibliographystyle{plain}
\bibliography{main.bib}

\newpage
\appendix

\counterwithin{table}{section}
\section{Appendix: Proofs}\label{proofs_sup}
\textbf{Lemma \ref{lemma1}. }
Let $\rmP_{\mathcal{A}}$ denote a matrix whose entry in row $a$ and column $k$ is $p(A=a|K=k)$ (i.e., the prior of group $a$ in client $k$). 
Then, given a solution to the minimax problem across clients
\begin{equation}
    h^*,\bm\lambda^* \in \arg\min\limits_{h \in \mathcal{H}}\max\limits_{\bm\lambda \in \Delta^{|\mathcal{K}|-1}} \mathop{\E}\limits_{\mathcal{D}_{\bm \lambda}}[\ell(h(X),Y)], 
\end{equation}
$\exists$ $\bm\mu^*=\rmP_{\mathcal{A}}\bm\lambda^*$ that is solution to the following constrained minimax problem across sensitive groups:
\begin{equation}
     h^*,\bm\mu^* \in \arg\min\limits_{h \in \mathcal{H}}\max\limits_{\bm \mu \in \rmP_{\mathcal{A}}\Delta^{|\mathcal{K}|-1}}\mathop{\E}\limits_{\mathcal{D}_{\bm \mu}}[\ell(h(X),Y)],
\end{equation}
where the weighting vector $\bm\mu$ is constrained to belong to the simplex subset defined by $\rmP_{\mathcal{A}}\Delta^{|\mathcal{K}|-1} \subseteq \Delta^{|\mathcal{A}|-1}$.
In particular, if the set ${\Gamma} = \big\{ \bm\mu' \in \rmP_{\mathcal{A}}\Delta^{|\mathcal{K}|-1}$: ${\bm\mu'} \in \arg\min\limits_{h \in \mathcal{H}}\max\limits_{ {\bm \mu} \in \Delta^{|\mathcal{A}|-1}}\mathop{\E}\limits_{\mathcal{D}_{\bm \mu}}[\ell(h(X),Y)]\big\} \not= \emptyset$, then ${\bm \mu}^* \in \Gamma $, and the minimax fairness solution across clients is also a minimax fairness solution across demographic groups. 

\bigskip

\noindent \textsc{Proof.}
The objective for optimizing the global model for the worst mixture of client distributions is:
\begin{equation}
\label{afl_objective}
\min\limits_{h \in \mathcal{H}} \max\limits_{\boldsymbol\lambda \in \Delta^{|\mathcal{K}|-1} }
\mathop{\E}\limits_{\mathcal{D}_{\boldsymbol\lambda}}[l(h(X),Y)] = \min\limits_{h \in \mathcal{H}} \max\limits_{\boldsymbol\lambda \in \Delta^{|\mathcal{K}|-1}} \sum\limits_{k=1}^{|\mathcal{K}|} \lambda_k\mathop{\E}\limits_{p(X,Y|K=k)}[l(h(X),Y)],
\end{equation}

\noindent given that $\mathcal{D}_{\bm\lambda}=\sum\limits_{k=1}^{|\mathcal{K}|} \lambda_k p(X,Y|K=k)$. 
 Since $p(X,Y|K=k)=\sum\limits_{a \in \mathcal{A}}p(A=a|K=k)p(X,Y|A)$ 
 with $p(A=a|K=k)$ being the prior of $a \in \mathcal{A}$ for client $k$, and $p(X,Y|A=a)$ is the distribution conditioned on the sensitive group $a \in \mathcal{A}$, Eq. \ref{afl_objective} can be re-written as
\begin{equation}
\label{clients_groups_equivalence}
\begin{array}{l}

    \min\limits_{h \in \mathcal{H}} \max\limits_{\boldsymbol\lambda \in \Delta^{|\mathcal{K}|-1}} \sum\limits_{k=1}^{|\mathcal{K}|} \lambda_k \sum\limits_{a \in \mathcal{A}} p(A=a|K=k)\underset{p(X,Y|A=a)}{\mathop{\E}}[l(h(X),Y)] = \\
    \\
    \min\limits_{h \in \mathcal{H}} \max\limits_{\boldsymbol\lambda \in \Delta^{|\mathcal{K}|-1}} \sum\limits_{a \in \mathcal{A}}\underset{p(X,Y|A=a)}{\mathop{\E}}[l(h(X),Y)] \Big ( \sum\limits_{k=1}^{|\mathcal{K}|} p(A=a|K=k)\lambda_k \Big ) = \\
    \\
    \min\limits_{h \in \mathcal{H}} \max\limits_{\boldsymbol\mu \in \rmP_\mathcal{A} \Delta^{|\mathcal{K}|-1}} \sum\limits_{a \in \mathcal{A}} \mu_a\underset{p(X,Y|A=a)}{\mathop{\E}}[l(h(X),Y)].
\end{array}
\end{equation}

where $\mu_a= \sum\limits_{k=1}^{|\mathcal{K}|} p(A=a|K=k)\lambda_k$, $\forall a \in \mathcal{A}$. Note that this creates the vector $\boldsymbol\mu=\rmP_\mathcal{A}\boldsymbol\lambda \subseteq \rmP_\mathcal{A} \Delta^{|\mathcal{K}|-1}$. It holds that the set of possible $\bm \mu$ vectors satisfies $ \rmP_\mathcal{A} \Delta^{|\mathcal{K}|-1} \subseteq \Delta^{|\mathcal{A}|-1}$, since $\rmP_{\mathcal{A}}= \big \{\{p(A=a|K=k)\}_{a \in \mathcal{A}} \big \}_{ k \in \mathcal{K}} \in  {\rm I\!R}_+^{|\mathcal{A}|\times |\mathcal{K}|}$, with $\underset{a \in \mathcal{A}}{\sum}p(A=a|K=k)=1$ $\forall k$ and $\boldsymbol\lambda \in \Delta^{|\mathcal{K}|-1}$. 

Then, from the equivalence in Equation \ref{clients_groups_equivalence} we have that
\begin{equation}
\label{eq:lambda_solution}
    h^*,\boldsymbol\lambda^* \in \arg\min\limits_{h \in \mathcal{H}}\max\limits_{\boldsymbol\lambda \in \Delta^{|\mathcal{K}|-1}} \mathop{\E}\limits_{\mathcal{D}_{\bm\lambda}}[\ell(h(X),Y)], 
\end{equation}
and
\begin{equation}
\label{eq:gamma_solution}
h^*,\boldsymbol\mu^* \in \arg\min\limits_{h \in \mathcal{H}}\max\limits_{\mu \in \rmP_{\mathcal{A}}\Delta^{|\mathcal{K}|-1}}\mathop{\E}\limits_{\mathcal{D}_{\bm\mu}}[\ell(h(X),Y)],
\end{equation}
with $\boldsymbol\mu^*=\rmP_{\mathcal{A}}\boldsymbol\lambda^*$ have the same minimax risk, that is
\begin{equation}
\mathop{\E}\limits_{\mathcal{D}_{\bm\mu^*}}[\ell(h^*(X),Y)] = \mathop{\E}\limits_{\mathcal{D}_{\bm\lambda^*}}[\ell(h^*(X),Y)].
\end{equation}

In particular, if the space $\rmP_{\mathcal{A}}\Delta^{|\mathcal{K}|-1}$ contains any  group minimax fair weights, meaning that the set ${\Gamma} = \big\{ \boldsymbol\mu' \in \rmP_{\mathcal{A}}\Delta^{|\mathcal{K}|-1}$: ${\boldsymbol\mu'} \in \arg\min\limits_{h \in \mathcal{H}}\max\limits_{{\bm\mu} \in \Delta^{|\mathcal{A}|-1}}\mathop{\E}\limits_{\mathcal{D}_{\bm\mu}}[\ell(h(X),Y)]\big\}$ is not empty, then it follows that any $\bm \mu^*$ (solution to Equation \ref{eq:gamma_solution}) is already minimax fair with respect to the groups ${\bm\mu}^* \in \Gamma$, and the client-level minimax solution is also a minimax solution across sensitive groups.




\label{lemma_1_proof}
\textbf{Lemma \ref{centralized_vs_federated_solution}. }
Consider our federated learning setting (Figure \ref{fig:Fairschemes}, right) where each entity $k$ has access to a local dataset $\mathcal{S}_k = \bigcup\limits_{a \in \mathcal{A}} \mathcal{S}_{a,k}$, and a centralized machine learning setting (Figure \ref{fig:Fairschemes}, left) where there is a single entity that has access to a single dataset $\mathcal{S} = \bigcup\limits_{k \in \mathcal{K}} \mathcal{S}_k = \bigcup\limits_{k \in \mathcal{K}} \bigcup\limits_{a \in \mathcal{A}} \mathcal{S}_{a,k}$ (i.e., this single entity in the centralized setting has access to the data of the various clients in the distributed setting). 
Then, Algorithm \ref{alg:fedminmax} (federated) and Algorithm \ref{alg:centralized_minmax} (non-federated, in supplementary material, Appendix \ref{centralized_algo}) lead to the same global model provided that learning rates and model initialization are identical.

\bigskip
 \noindent {\textsc{Proof.}} We will show that FedMinMax, in Algorithm \ref{alg:fedminmax} is equivalent to the centralized algorithm, in Algorithm \ref{alg:centralized_minmax} under the following conditions:
 \begin{enumerate}

     \item the dataset on client $k$, in FedMinMax is $\mathcal{S}_k = \bigcup\limits_{a \in \mathcal{A}} \mathcal{S}_{a,k}$ and the dataset in centralized MinMax is $\mathcal{S} = \bigcup\limits_{k \in \mathcal{K}} \mathcal{S}_k = \bigcup\limits_{k \in \mathcal{K}} \bigcup\limits_{a \in \mathcal{A}} \mathcal{S}_{a,k}$; and
     
    \item the model initialization $\theta^0$, the number of adversarial rounds $T$,\footnote{In the federated Algorithm \ref{alg:fedminmax}, we also refer to the adversarial rounds as communication rounds.} learning rate for the adversary $\eta_\mu$, and learning rate for the learner $\eta_\theta$, are identical for both algorithms.
 \end{enumerate}

This can then be immediately done by showing that steps lines 3-7 in Algorithm 1 are entirely equivalent to step 3 in Algorithm \ref{alg:centralized_minmax}. In particular, note that we can write
\begin{equation*}
\begin{array}{ll}
\hat{r}(\theta, \bm \mu) & =\sum\limits_{a \in \mathcal{A}}\mu_a\hat{r}_a(\bm \theta)\\

\\&= \sum\limits_{a \in \mathcal{A}}\mu_a \sum\limits_{k \in \mathcal{K}}\frac{n_{a,k}}{n_a}\hat{r}_{a,k}(\bm \theta)\\

\\&= \sum\limits_{a \in \mathcal{A}}\mu_a\frac{n}{n_a}\frac{1}{n} \sum\limits_{k \in \mathcal{K}}n_{a,k}\hat{r}_{a,k}(\bm \theta)
\\

\\&= \sum\limits_{a \in \mathcal{A}} w_a \frac{1}{n} \sum\limits_{k \in \mathcal{K}} \frac{n_{a,k}}{n_k} n_k \hat{r}_{a,k}(\bm \theta)
\\

\end{array}
\end{equation*}

\begin{equation}
\begin{array}{ll}\label{equivalence}
\\&= \sum\limits_{k \in \mathcal{K}} \frac{n_k}{n} \sum\limits_{a \in \mathcal{A}} w_a \frac{n_{a,k}}{n_k} \hat{r}_{a,k}(\bm \theta)
\\
\\&= \sum\limits_{k \in \mathcal{K}} \frac{n_k}{n} \hat{r}_{k}(\bm \theta, \bm w),
\\
\end{array}
\end{equation}

\begin{equation}\label{grouprisk}
\textnormal{where } \hat{r}_k(\bm \theta, \bm w) = \sum\limits_{a \in \mathcal{A}}\frac{n_{a,k}}{n_k} w_a \hat{r}_{a,k}(\bm \theta)\textnormal{, with } w_a= \frac{\mu_a}{\frac{n_a}{n}}, \textnormal{ and }
\hat{r}_a (\bm \theta) = \sum\limits_{k \in \mathcal{K}} \frac{n_{a,k}}{n_a} \hat{r}_{a,k}(\bm \theta).
\end{equation}

Therefore, the model update 
\begin{equation}
    \bm \theta^{t}= \sum\limits_{k \in \mathcal{K}} \frac{n_k}{n} \bm \theta^{t}_k  =  \sum\limits_{k \in \mathcal{K}} \frac{n_k}{n}  \big(\bm \theta^{t-1} - \eta_\theta \nabla_\theta \hat{r}_k(\bm \theta^{t-1}, \bm w^{t-1}) \big)
\end{equation}
associated with step in 7 at round $t$ of Algorithm 1, is entirely equivalent to the model update
\begin{equation}
    \bm \theta^{t} = \bm \theta^{t-1} - \eta_\theta \nabla_\theta \hat{r}(\bm \theta^{t-1}, \bm w^{t-1})
\end{equation}
associated with step in line 3 at round $t$ of Algorithm \ref{alg:centralized_minmax}, provided that ${\bm \theta}^{t-1}$ is the same for both algorithms. 

It follows therefore by induction that, provided the initialization ${\bm \theta}^{0}$ and learning rate $\eta_\theta$ are identical in both cases the algorithms lead to the same model.
Also, from Eq. \ref{grouprisk}, we have that the projected gradient ascent step in line 4 of Algorithm \ref{alg:centralized_minmax} is equivalent to the step in line 10 of Algorithm \ref{alg:fedminmax}.
\label{lemma_2_proof}

\section{Appendix: Centralized Minimax Algorithm}\label{centralized_algo}

 We provide the centralized version of FedMinMax in Algorithm \ref{alg:centralized_minmax}.

\begin{algorithm}[ht]

    \caption{\textsc{Centralized MinMax Baseline}}
    \label{alg:centralized_minmax}
    {\bfseries Input:} $T:$ total number of adversarial rounds,
     $\eta_{\bm \theta}$: model learning rate, $\eta_{\bm\mu}$: adversary learning rate, $\mathcal{S}_{a}$: set of examples for group $a$, $\forall a \in \mathcal{A}$. 
    \begin{algorithmic}[1]
    \setstretch{1.35}
    
    \STATE Server {\bfseries initializes} $\bm\mu^0\leftarrow \{|\mathcal{S}_a|/|\mathcal{S}| \}_{a \in \mathcal{A}}$ and $\bm \theta^0$ randomly.
    
    \FOR{$t=1$ {\bfseries to} $T$}

    \STATE Server {\bfseries computes}  $\bm \theta^{t}_k \leftarrow \bm \theta^{t-1} - \eta_\theta \nabla_\theta \hat{r}(\bm \theta^{t-1}, \bm \mu^{t-1})$ 
    
     \STATE Server {\bfseries updates}
     \newline
     $\bm\mu^{t} \leftarrow \prod_{\Delta^{|\mathcal{A}|-1}} \Big (\bm{{\mu}}^{t-1} +\eta_{\bm{\mu}} {\nabla}_{\bm{\mu}}\langle\ \bm \mu^{t-1},\hat{r}_a(\bm \theta^{t-1})\rangle \Big ) $
     \ENDFOR
    
    \end{algorithmic}
    \noindent{\bfseries Outputs}: $\frac{1}{T} \sum_{t =1}^T \bm \theta^t$
    \end{algorithm}

\section{Appendix: Experimental Details}\label{more_results}

\noindent \textbf{Experimental Setting and Model Architectures.} For AFL and FedMinMax the batch size is equal to the number of examples per client while for TERM, FedAvg and $q$-FedAvg is equal to $100$.  
For the synthetic dataset, we use an MLP architecture consisting of four hidden layers of size 512. In the experiments for Adult we use a single layer MLP with 512 neurons.
For FashionMNIST we use a CNN architecture with two 2D convolutional layers with kernel size 3, stride 1, and padding 1. Each convolutional layer is followed with a maxpooling layer with kernel size 2, stride 2, dilation 1, and padding 0.  For CIFAR-10 we use a ResNet-18 architecture without batch normalization. Finally for ACS Employment dataset we use a single layer MLP with 512 neurons for the experiments where the sensitive label is the combination of race and employment, and Logistic Regression for the experiments with the original 9 races.  For training we use either cross entropy or Brier score loss function. We perform a grid search over the following hyperparameters: tilt-$t=\{0.01,0.1,0.5,0.8,1.0\}$, $q=\{0.2,0.5,1.0,2.0,5.0\}$, local epochs $E=\{3,10,15\}$ and $\eta_\theta=\eta_\mu=\eta_\lambda =\{0.001,0.005,0.01,0.05,0.1\}$ (where appropriate).
We report a summary of the experimental setup in Table \ref{tab:training_settings}. 
During the training process we tune the hyperparameters based on the validation set for each approach. The mean and standard deviation reported on the results are calculated over three runs. We use 3-fold cross validation to split the data into training and validation for each run.

\begin{table}[H]
\centering
\caption{Summary of parameters used in the training process for all experiments. Epochs refer to the local iterations performed at each client, $n_k$ is the number of local data examples in client $k$, $\eta_\theta$ is the model's learning rate and $\eta_\mu$ or $\eta_\lambda$ is the adversary learning rates.}
\resizebox{0.95\textwidth}{!}{
\begin{tabular}{lcccccccc}
\toprule
\textbf{Dataset}&{\textbf{Setting} }  &  \textbf{Method} &  $\eta_\theta$  &  \textbf{Batch Size} &  \textbf{Loss} & {\textbf{Hypothesis Type}} & \textbf{Epochs} & \textbf{$\eta_\mu$ or $\eta_\lambda$}  \\
\midrule
Synthetic&{ESG,SSG}   & AFL &  0.1& $n_k$& Brier Score & MLP (4x512)&-&0.1 \\
& &FedAvg & 0.1  & 100  & Brier Score & MLP (4x512)&15& - \\
& &$q$-FedAvg & 0.1  & 100  & Brier Score & MLP (4x512)&15& -  \\
& &FedMinMax (ours) & 0.1 & $n_k$  & Brier Score &MLP (4x512)&- &0.1 \\
& &{Centalized Minmax} & 0.1&$n$ &Brier Score &MLP (4x512) &-& 0.1 \\
\midrule
Adult &  {ESG,SSG,PSG } & AFL &  0.01& $n_k$& Cross Entropy & MLP (512)&-&0.01 \\
&&FedAvg & 0.01  & 100  & Cross Entropy & MLP (512)&15& - \\
& &$q$-FedAvg & 0.01  & 100  & Cross Entropy & MLP (512)&15& -  \\
& &FedMinMax (ours) & 0.01 & $n_k$  & Cross Entropy &MLP (512)&- &0.01 \\
& &{Centalized Minmax} & 0.01&$n$ &Cross Entropy &MLP (512) &-& 0.01 \\
\midrule
FashionMNIST&{ESG,SSG,PSG }& AFL &0.1 & $n_k$& Brier Score&  CNN&- & 0.1\\
& &FedAvg &  0.1 & 100 & Brier Score  & CNN&15 & -\\
& &$q$-FedAvg  &  0.1 & 100 & Brier Score  & CNN&15 & -\\
& &FedMinMax (ours) & 0.1 &  $n_k$ &Brier Score&CNN&-& 0.1\\
& &{Centalized Minmax} & 0.1& $n$&Brier Score&CNN&-&0.1 \\
\midrule
CIFAR-10&{ESG,SSG,PSG }& AFL &0.1 & $n_k$& Brier Score&  ResNet-18 w/o BN&- & 0.01\\
& &FedAvg &  0.1 & 100 & Brier Score  & ResNet-18 w/o BN & 3 & -\\
& &$q$-FedAvg &  0.1 & 100 & Brier Score  & ResNet-18 w/o BN & 3 & -\\
& &FedMinMax (ours) & 0.1 &  $n_k$ &Brier Score&ResNet-18 w/o BN&-& 0.01\\
& &{Centalized Minmax} & 0.1& $n$&Brier Score&ResNet-18 w/o BN&-&0.01 \\
\midrule
ACS Employment&{ESG,SSG,PSG }& AFL &0.01 & $n_k$& Cross Entropy&   MLP (512)&- & 0.01\\
(6 sensitive groups)& &FedAvg &  0.01 & 100 & Cross Entropy  &  MLP (512) & 10 & -\\
& &$q$-FedAvg &  0.01 & 100 & Cross Entropy  &  MLP (512) & 10 & -\\
& &FedMinMax (ours) & 0.01 &  $n_k$ &Cross Entropy& MLP (512)&-& 0.01\\
& &{Centalized Minmax} & 0.01& $n$&Cross Entropy& MLP (512)&-&0.01 \\
\midrule
ACS Employment&{ESG,SSG,PSG }& AFL &0.01 & $n_k$& Cross Entropy&  Logistic Regression&- & 0.01\\
(9 sensitive groups)& &FedAvg &  0.01 & 100 & Cross Entropy  & Logistic Regression & 10 & -\\
& &$q$-FedAvg &  0.01 & 100 & Cross Entropy  & Logistic Regression & 10 & -\\
& &FedMinMax (ours) & 0.01 &  $n_k$ &Cross Entropy&Logistic Regression&-& 0.01\\
& &{Centalized Minmax} & 0.01& $n$&Cross Entropy&Logistic Regression&-&0.01 \\
\bottomrule
\end{tabular}
}

\label{tab:training_settings}
\end{table}

\noindent\textbf{Software \& Hardware.} The proposed algorithms and experiments are written in Python, leveraging PyTorch \cite{NEURIPS2019_9015}. The experiments were realised using 1 $\times$ NVIDIA Tesla V100 GPU.

\bigskip

\section{Appendix: Additional Results}\label{additional_results}

\noindent\textbf{Experiments on Synthetic dataset.}
Recall that we consider two sensitive groups (i.e., $|\mathcal{A}|=2$) in the synthetic dataset. 
In the \textit{Equal access to Sensitive Groups (ESG)} setting, we distribute the two groups on 40 clients, while for the \textit{Single access to Sensitive Groups (SSG)} case, every client has access to a single group, each group is distributed to 20 clients, and the amount of samples on each local dataset varies across clients.  
There is no \textit{Partial access to Sensitive Groups (PSG)} setting for 
binary sensitive group scenarios since it is equivalent to SSG. A comparison of the testing group risks is provided in Table \ref{tab:synthetic} and the weighting coefficients for the groups are given by Table \ref{tab:coeff_synthetic}.

\begin{table}[ht]

\centering
\caption{Final group weighting coefficients for AFL and FedMinmax across different federated learning scenarios on the synthetic dataset for binary classification involving two sensitive groups.}

\resizebox{0.45\textwidth}{!}{
\begin{tabular}{llll}
\toprule
 \textbf{Setting}&  \textbf{Method} & \textbf{Worst Group}  & \textbf{Best Group}\\
\midrule
 ESG & AFL & 0.528 & 0.472 \\
& FedMinMax (ours) & 0.999 & 0.001 \\

\midrule
 SSG & AFL & 0.999 &  0.001 \\
& FedMinMax (ours) & 0.999 &  0.001 \\

\midrule
\multicolumn{2}{c}{Centalized Minmax Baseline} & 0.999 &  0.001 \\
\bottomrule
\end{tabular}}

\label{tab:coeff_synthetic}
\end{table}

\begin{table}[ht]
\centering
\caption{Testing Brier score risks for FedAvg, AFL, $q$-FedAvg, TERM, and FedMinmax across different federated learning scenarios on the synthetic dataset for binary classification involving two sensitive groups. PSG scenario is not included because for $|\mathcal{A}|=2$ it is equivalent to SSG.}

\resizebox{0.5\textwidth}{!}{
\begin{tabular}{llll}
\toprule
 \textbf{Setting}&  \textbf{Method} &     \textbf{Worst Group Risk}  & \textbf{Best Group Risk}\\
\midrule
 ESG & AFL &    0.485$\pm$0.0 &  0.216$\pm$0.001  \\
&FedAvg &    0.487$\pm$0.0 &  0.214$\pm$0.002  \\
& $q$-FedAvg ($q$=0.2) &  0.479$\pm$0.002 &   0.22$\pm$0.002  \\
& $q$-FedAvg ($q$=5.0) &  0.478$\pm$0.002 &  0.223$\pm$0.004  \\
& TERM ($t$=1.0) &  0.469$\pm$0.0 &  0.261$\pm$0.001  \\
& FedMinMax (ours) &    \textbf{0.451$\pm$0.0} &   \textbf{0.31$\pm$0.001} \\

\midrule
 SSG & AFL &    \textbf{0.451$\pm$0.0} & \textbf{0.31$\pm$0.001} \\
& FedAvg &  0.483$\pm$0.002 &  0.219$\pm$0.001  \\
& $q$-FedAvg ($q$=0.2) &   0.476$\pm$0.001 &  0.221$\pm$0.002  \\
& $q$-FedAvg ($q$=5.0) &  0.468$\pm$0.005 &  0.274$\pm$0.004  \\
& TERM ($t$=1.0) &  0.461$\pm$0.004 &  0.272$\pm$0.001  \\
& FedMinMax (ours) &    \textbf{0.451$\pm$0.0} &  \textbf{0.309$\pm$0.003} \\

\midrule
\multicolumn{2}{c}{Centalized Minmax Baseline} &
 \textbf{0.451$\pm$0.0} &  \textbf{0.308$\pm$0.001} \\
\bottomrule
\end{tabular}
}

\label{tab:synthetic}
\end{table}

\noindent\textbf{Experiments on Adult dataset.} 
In the \textit{Equal access to Sensitive Groups (ESG)} setting, we distribute the 4 groups equally on 40 clients.
In the \textit{Partial access to Sensitive Groups (PSG)} setting, 20 clients have access to \textit{Males} subgroups, and the other 20 to subgroups relating to \textit{Females}. In the \textit{Single access to Sensitive Groups (SSG)} setting, every client has access to a single group and each group is distributed to 10 clients. We show the testing group risks in Table \ref{tab:analytic_adult_risk} and the group weights in Table \ref{tab:coeff_adult}.

\begin{table}[ht]
\centering
\caption{Final group weighting coefficients for AFL and FedMinmax across different federated learning scenarios on the Adult dataset. We round the weights values to the last three decimal places.}
\resizebox{0.89\textwidth}{!}{
\begin{tabular}{llcccc}
\toprule
 \textbf{Setting}&  \textbf{Method} & \textbf{Males earning <= $50$K}  & \textbf{Males earning > $50$K} & \textbf{Females earning <= $50$K}  & \textbf{Females earning > $50$K}  \\
\midrule
 ESG & AFL &  0.475 & 0.214 &    0.284 &   0.028   \\
& FedMinMax (ours) &   0.697 & 0.301 & 0.001 & 0.001 \\

\midrule
 SSG & AFL &  0.705 & 0.293 & 0.003 & 0.001 \\
& FedMinMax (ours) & 0.697 & 0.301 & 0.001 & 0.001 \\

\midrule
 PSG & AFL & 0.500 & 0.229 & 0.244 & 0.027 \\
& FedMinMax (ours) & 0.705 & 0.293 & 0.001 & 0.001 \\
\midrule
\multicolumn{2}{c}{Centalized Minmax Baseline} & 0.697 & 0.301 & 0.001 &   0.001 \\
\bottomrule
\end{tabular}}

\label{tab:coeff_adult}
\end{table}

\begin{table}[H]
\centering
\caption{Cross entropy risks for FedAvg, AFL, q-FedAvg, TERM, and FedMinmax across different federated learning settings on adult dataset.}
\resizebox{0.89\textwidth}{!}{
\begin{tabular}{llcccc}
\toprule
 \textbf{Setting}&  \textbf{Method} & \textbf{Males earning <= $50$K}  & \textbf{Males earning > $50$K} & \textbf{Females earning <= $50$K}  & \textbf{Females earning > $50$K}  \\
\midrule
 ESG & AFL &  0.263$\pm$0.002 &  0.701$\pm$0.003 &  0.086$\pm$0.002 &  1.096$\pm$0.008 \\
&FedAvg &  0.255$\pm$0.002 &  0.697$\pm$0.004 &  0.081$\pm$0.001 &  1.121$\pm$0.009 \\
&q-FedAvg &  0.263$\pm$0.003 &  0.697$\pm$0.004 &  0.084$\pm$0.001 &    1.1$\pm$0.006  \\
& TERM &  0.381$\pm$0.101 &  0.607$\pm$0.04 &  0.224$\pm$0.06 &  0.725$\pm$0.021 \\
&FedMinMax (ours) &  0.414$\pm$0.003 &  0.453$\pm$0.003 &  0.415$\pm$0.008 &  \textbf{0.347$\pm$0.007} \\
\midrule
SSG &AFL &  0.418$\pm$0.006 &  0.452$\pm$0.009 &  0.416$\pm$0.002 &  \textbf{0.349$\pm$0.007}   \\
&FedAvg &  0.263$\pm$0.001 &  0.704$\pm$0.002 &     0.07$\pm$0.0 &   1.23$\pm$0.002    \\
&q-FedAvg &  0.261$\pm$0.001 &  0.683$\pm$0.002 &  0.082$\pm$0.001 &   1.117$\pm$0.01   \\
&TERM &  0.358$\pm$0.016 &  0.579$\pm$0.002 &  0.286$\pm$0.031 &  0.693$\pm$0.071  \\
&FedMinMax (ours) &  0.413$\pm$0.002 &  0.453$\pm$0.005 &  0.414$\pm$0.006 &   \textbf{0.348$\pm$0.01}  \\

\midrule
PSG & AFL &  0.274$\pm$0.003 &  0.757$\pm$0.009 &  0.094$\pm$0.002 &  1.285$\pm$0.022 \\
&FedAvg &  0.263$\pm$0.001 &    0.7$\pm$0.001 &  0.069$\pm$0.001 &  1.226$\pm$0.007  \\
&q-FedAvg &  0.263$\pm$0.004 &  0.752$\pm$0.014 &   0.09$\pm$0.004 &  1.239$\pm$0.032 \\
&TERM &   0.485$\pm$0.195 &  0.581$\pm$0.108 &  0.367$\pm$0.316 &  0.69$\pm$0.003  \\
&FedMinMax (ours) &  0.411$\pm$0.002 &  0.452$\pm$0.006 &  0.417$\pm$0.001 &  \textbf{0.346$\pm$0.008}\\
\midrule
\multicolumn{2}{c}{Centalized Minmax Baseline} & 0.412$\pm$0.004 &  0.453$\pm$0.005 &  0.416$\pm$0.012 &  \textbf{0.347$\pm$0.004} \\

\bottomrule
\end{tabular}}
\label{tab:analytic_adult_risk} 
\end{table}

\bigskip

\noindent \textbf{Experiments on FashionMNIST dataset.}
 For the \textit{Equal access to Sensitive Groups (ESG)} setting, each client in the federation has access to the same amount of the 10 classes.
 In the \textit{Partial access to Sensitive Groups (PSG)} setting, 20 of the participants have access only to groups \textit{T-shirt}, \textit{Trouser}, \textit{Pullover}, \textit{Dress} and \textit{Coat}. The remaining 20 clients own data from groups \textit{Sandal}, \textit{Shirt}, \textit{Sneaker}, \textit{Bag} and \textit{Ankle Boot}. 
  Finally, in the \textit{Single access to Sensitive Groups (SSG)} setting, every group is owned by 4 clients only and all clients have access to just one group membership. 
  The group risks are provided in Table \ref{tab:analytic_fmnist}. We also show the weighting coefficients for each sensitive group in Table \ref{tab:coeff_fmnist}.

\begin{table}[ht]
\caption{Brier score risks for FedAvg, AFL, q-FedAvg, TERM, and FedMinmax across different federated learning settings on FashionMNIST dataset.
}
\resizebox{\textwidth}{!}{
\begin{tabular}{llcccccccccc}
\toprule
 \textbf{Setting}&  \textbf{Method} &\textbf{T-shirt} & \textbf{Trouser} & \textbf{Pullover} & \textbf{Dress} & \textbf{Coat} & \textbf{Sandal} & \textbf{Shirt} & \textbf{Sneaker} & \textbf{Bag} & \textbf{Ankle boot}\\
\midrule
 ESG &AFL &  0.239$\pm$0.003 &    0.046$\pm$0.0 &  0.262$\pm$0.001 &  0.159$\pm$0.001 &  0.252$\pm$0.004 &     0.06$\pm$0.0 &  0.494$\pm$0.004 &  0.067$\pm$0.001 &    0.049$\pm$0.0 &   0.07$\pm$0.001 \\
&FedAvg &  0.243$\pm$0.003 &    0.046$\pm$0.0 &  0.262$\pm$0.001 &  0.158$\pm$0.003 &  0.253$\pm$0.002 &    0.061$\pm$0.0 &  0.492$\pm$0.003 &    0.068$\pm$0.0 &    0.049$\pm$0.0 &    0.069$\pm$0.0 \\
&q-FedAvg &  0.268$\pm$0.051 &  0.047$\pm$0.005 &  0.312$\pm$0.016 &  0.164$\pm$0.029 &  0.306$\pm$0.052 &  0.039$\pm$0.003 &  0.477$\pm$0.006 &  0.074$\pm$0.001 &  0.036$\pm$0.005 &  0.056$\pm$0.008 \\
&TERM &  0.256$\pm$0.066 &  0.048$\pm$0.008 &   0.31$\pm$0.083 &  0.175$\pm$0.022 &  0.294$\pm$0.016 &  0.041$\pm$0.012 &  0.467$\pm$0.002 &  0.066$\pm$0.019 &  0.038$\pm$0.011 &  0.062$\pm$0.018 \\
&FedMinMax (ours) &  0.261$\pm$0.006 &  0.191$\pm$0.016 &  0.256$\pm$0.027 &  0.217$\pm$0.013 &  0.223$\pm$0.031 &  0.207$\pm$0.027 &   \textbf{0.307$\pm$0.01} &  0.172$\pm$0.016 &  0.193$\pm$0.021 &  0.156$\pm$0.011 \\
\midrule
SSG &AFL &  0.267$\pm$0.009 &  0.194$\pm$0.023 &  0.236$\pm$0.013 &  0.226$\pm$0.012 &  0.262$\pm$0.012 &  0.201$\pm$0.026 &  \textbf{0.307$\pm$0.003} &  0.178$\pm$0.033 &  0.205$\pm$0.025 &  0.162$\pm$0.021 \\
&FedAvg &  0.227$\pm$0.003 &  0.039$\pm$0.001 &  0.236$\pm$0.004 &  0.143$\pm$0.003 &  0.232$\pm$0.003 &  0.051$\pm$0.001 &  0.463$\pm$0.003 &    0.067$\pm$0.0 &    0.041$\pm$0.0 &  0.063$\pm$0.001 \\
 
&q-FedAvg &  0.24$\pm$0.001 &  0.041$\pm$0.008 &  0.246$\pm$0.026 &  0.142$\pm$0.014 &  0.257$\pm$0.028 &  0.036$\pm$0.001 &  0.425$\pm$0.002 &  0.059$\pm$0.014 &  0.027$\pm$0.002 &  0.042$\pm$0.007 \\ 
&TERM &0.251$\pm$0.011 &  0.034$\pm$0.003 &   0.26$\pm$0.017 &  0.144$\pm$0.005 &  0.242$\pm$0.034 &   0.04$\pm$0.004 &  0.399$\pm$0.017 &   0.05$\pm$0.003 &  0.026$\pm$0.001 &  0.044$\pm$0.001 \\

&FedMinMax (ours) &  0.269$\pm$0.012 &    0.2$\pm$0.026 &  0.238$\pm$0.017 &  0.231$\pm$0.013 &  0.252$\pm$0.034 &    0.2$\pm$0.024 &  \textbf{0.309$\pm$0.011} &   0.177$\pm$0.03 &  0.205$\pm$0.032 &  0.169$\pm$0.013 \\
\midrule
PSG & AFL &  0.244$\pm$0.007 &  0.032$\pm$0.001 &  0.257$\pm$0.066 &  0.122$\pm$0.006 &  0.209$\pm$0.098 &  0.045$\pm$0.002 &  0.425$\pm$0.019 &  0.059$\pm$0.001 &  0.041$\pm$0.001 &  0.062$\pm$0.001 \\
&FedAvg &  0.229$\pm$0.008 &    0.039$\pm$0.0 &  0.236$\pm$0.004 &  0.142$\pm$0.002 &  0.232$\pm$0.003 &  0.052$\pm$0.001 &  0.464$\pm$0.011 &  0.067$\pm$0.001 &  0.042$\pm$0.001 &  0.063$\pm$0.001 \\
&q-FedAvg & 0.278$\pm$0.062 &   0.04$\pm$0.013 &  0.256$\pm$0.083 &   0.16$\pm$0.026 &  0.311$\pm$0.044 &  0.045$\pm$0.013 &  0.453$\pm$0.002 &   0.063$\pm$0.02 &  0.029$\pm$0.007 &  0.047$\pm$0.004 \\       
&TERM & 0.226$\pm$0.007 &  0.037$\pm$0.005 &  0.233$\pm$0.004 &  0.153$\pm$0.007 &  0.255$\pm$0.016 &    0.038$\pm$0.0 &  0.439$\pm$0.007 &  0.053$\pm$0.003 &  0.026$\pm$0.001 &  0.043$\pm$0.002 \\
&FedMinMax (ours) &  0.263$\pm$0.013 &  0.177$\pm$0.026 &  0.228$\pm$0.011 &   0.21$\pm$0.019 &  0.238$\pm$0.025 &   0.182$\pm$0.03 &   \textbf{0.31$\pm$0.008} &   0.16$\pm$0.027 &  0.184$\pm$0.031 &  0.154$\pm$0.018 \\
\midrule
\multicolumn{2}{c}{Centalized Minmax Baseline} &  0.259$\pm$0.01 &  0.173$\pm$0.015 &  0.239$\pm$0.051 &  0.213$\pm$0.008 &  0.24$\pm$0.063 &  0.182$\pm$0.024 & \textbf{0.311$\pm$0.006} &  0.168$\pm$0.018 &  0.18$\pm$0.013 &  0.151$\pm$0.012 \\

\bottomrule
\end{tabular}}

\label{tab:analytic_fmnist} 
\end{table}

\begin{table}[H]
\centering
\caption{Final group weighting coefficients for AFL, Centalized Minmax Baseline, and FedMinmax across different federated learning scenarios on the FashionMNIST dataset. Note that the weighting coefficients are rounded to the last three decimal places. We highlight the weighting coefficient for the worst group.
}
\resizebox{0.8\textwidth}{!}{
\begin{tabular}{llcccccccccc}
\toprule
 \textbf{Setting}&  \textbf{Method} &\textbf{T-shirt}  & \textbf{Trouser} & \textbf{Pullover} & \textbf{Dress} & \textbf{Coat} & \textbf{Sandal} & \textbf{Shirt} & \textbf{Sneaker} & \textbf{Bag} & \textbf{Ankle boot}\\
\midrule
 ESG &AFL &  0.099 &    0.100 &     0.101 &  0.101 &  0.100 &   0.100 &  0.099 &    0.100 &  0.100 & 0.100  \\
&FedMinMax (ours) & 0.217 & 0.001 &     0.241 &  0.007 &  0.151 &   0.001 &  \textbf{0.380} &    0.001 &  0.001 & 0.001\\
\midrule
SSG &AFL &   0.217 &    0.001 &     0.241 &  0.007 &  0.151 &   0.001 &  \textbf{0.379} &    0.001 &  0.001 & 0.001 \\
&FedMinMax (ours) &   0.216 &    0.001 &     0.237 &  0.017 &  0.155 &   0.001 &  \textbf{0.370} &    0.001 &  0.001 & 0.001  \\
\midrule
PSG & AFL &   0.128 &    0.064 &     0.138 &  0.099 &  0.129 &   0.063 &  0.173 &    0.069 &  0.066 & 0.071 \\
&FedMinMax (ours) &    0.216 &    0.001 &     0.238 &  0.014 &  0.154 &   0.001 &  \textbf{0.372}&    0.001 &  0.001 & 0.001\\
\midrule
\multicolumn{2}{c}{Centalized Minmax Baseline} & 0.217 &    0.001 &     0.240 &  0.010 &  0.152 &   0.001 &  \textbf{0.377} &    0.001 &  0.001 & 0.001 \\

\bottomrule
\end{tabular}}

\label{tab:coeff_fmnist} 
\end{table}

\bigskip

\noindent \textbf{Experiments on CIFAR-10 dataset.}
 In the \textit{Equal access to Sensitive Groups (ESG)} setting, the 10 classes are equally distributed across the clients, creating a scenario where each client has access to the same amount of data examples and groups.
 In the \textit{Partial access to Sensitive Groups (PSG)} setting,  20 clients own data from groups \textit{Airplane}, \textit{Automobile}, \textit{Bird}, \textit{Cat} and \textit{Deer} and the rest hold data from \textit{Dog}, \textit{Frog}, \textit{Horse}, \textit{Ship} and \textit{Truck} groups. Finally, in the \textit{Single access to Sensitive Groups (SSG)} setting, every client owns only one sensitive group and each group is distributed to only 4 clients. 
 We report the risks on the test set in Table \ref{analytic_cifar10} and the final group weighting coefficients in Table \ref{tab:coeff_cifar}.

\begin{table}[H]
\centering
\caption{Brier score risks for FedAvg, AFL, q-FedAvg, TERM, and FedMinmax across different federated learning settings on CIFAR-10 dataset.}
\resizebox{\textwidth}{!}{
\begin{tabular}{llcccccccccc}
\toprule
 \textbf{Setting}&  \textbf{Method} &\textbf{Airplane} &  \textbf{Automobile} &      \textbf{Bird} &       \textbf{Cat} &      \textbf{Deer} &       \textbf{Dog} &      \textbf{Frog} &     \textbf{Horse} &      \textbf{Ship} & \textbf{Truck}\\
 \midrule
ESG &AFL &   0.14$\pm$0.001 &  0.104$\pm$0.009 &  0.289$\pm$0.011 &   0.461$\pm$0.01 &   0.243$\pm$0.01 &   0.28$\pm$0.016 &  0.151$\pm$0.009 &   0.14$\pm$0.009 &  0.125$\pm$0.012 &  0.132$\pm$0.009 \\
&FedAvg &  0.148$\pm$0.014 &  0.108$\pm$0.006 &  0.283$\pm$0.011 &  0.487$\pm$0.002 &  0.237$\pm$0.002 &  0.256$\pm$0.002 &  0.144$\pm$0.005 &  0.148$\pm$0.008 &  0.123$\pm$0.003 &  0.128$\pm$0.004 \\
&q-FedAvg &0.178$\pm$0.065 &  0.118$\pm$0.047 &  0.308$\pm$0.099 &  0.507$\pm$0.003 &  0.311$\pm$0.054 &    0.41$\pm$0.01 &  0.179$\pm$0.012 &  0.119$\pm$0.013 &   0.158$\pm$0.07 &   0.182$\pm$0.05 \\
&TERM &0.217$\pm$0.087 &  0.115$\pm$0.006 &  0.311$\pm$0.057 &  0.491$\pm$0.007 &  0.274$\pm$0.055 &  0.272$\pm$0.026 &  0.176$\pm$0.041 &  0.166$\pm$0.013 &  0.175$\pm$0.069 &   0.12$\pm$0.006 \\

&FedMinMax (ours) &  0.257$\pm$0.003 &  0.189$\pm$0.009 &  0.324$\pm$0.015 &  \textbf{0.351$\pm$0.002} &  0.291$\pm$0.004 &   0.291$\pm$0.03 &  0.231$\pm$0.007 &  0.309$\pm$0.008 &  0.194$\pm$0.002 &  0.158$\pm$0.008  \\
\midrule
SSG &AFL &  0.283$\pm$0.027 &  0.259$\pm$0.001 &   0.18$\pm$0.008 &    \textbf{0.352$\pm$0.0} &  0.285$\pm$0.002 &  0.328$\pm$0.008 &  0.231$\pm$0.043 &  0.212$\pm$0.031 &  0.198$\pm$0.012 &  0.159$\pm$0.007 \\
& FedAvg &  0.189$\pm$0.011 &  0.102$\pm$0.009 &  0.253$\pm$0.005 &  0.485$\pm$0.017 &  0.239$\pm$0.079 &  0.339$\pm$0.074 &  0.148$\pm$0.021 &  0.166$\pm$0.029 &  0.121$\pm$0.019 &  0.138$\pm$0.022  \\
&q-FedAvg &0.18$\pm$0.026 &   0.11$\pm$0.017 &   0.29$\pm$0.016 &  0.437$\pm$0.002 &  0.334$\pm$0.069 &  0.345$\pm$0.009 &   0.161$\pm$0.03 &  0.175$\pm$0.057 &  0.176$\pm$0.105 &  0.129$\pm$0.013 \\
&TERM &0.149$\pm$0.015 &  0.146$\pm$0.014 &  0.378$\pm$0.042 &  0.392$\pm$0.021 &  0.262$\pm$0.039 &   0.307$\pm$0.02 &  0.192$\pm$0.052 &  0.176$\pm$0.003 &  0.167$\pm$0.032 &  0.119$\pm$0.029 \\

&FedMinMax (ours) &   0.258$\pm$0.01 &  0.187$\pm$0.005 &  0.332$\pm$0.005 &  \textbf{0.351$\pm$0.002} &  0.293$\pm$0.007 &  0.334$\pm$0.017 &  0.216$\pm$0.009 &  0.305$\pm$0.009 &  0.205$\pm$0.002 &  0.154$\pm$0.005  \\
\midrule
PSG & AFL &  0.158$\pm$0.019 &   0.121$\pm$0.01 &  0.289$\pm$0.015 &  0.439$\pm$0.006 &   0.247$\pm$0.01 &   0.28$\pm$0.014 &  0.151$\pm$0.016 &  0.168$\pm$0.011 &  0.125$\pm$0.013 &  0.118$\pm$0.009  \\
& FedAvg &  {0.167$\pm$0.005} &  {0.098$\pm$0.004 }&   {0.32$\pm$0.009} &  {0.471$\pm$0.014} &  {0.224$\pm$0.036} &  {0.304$\pm$0.009} &   {0.15$\pm$0.009} &  {0.162$\pm$0.028} &  {0.113$\pm$0.003} &  {0.121$\pm$0.013} \\
&q-FedAvg &  0.173$\pm$0.008 &  0.132$\pm$0.027 &  0.303$\pm$0.001 &   0.46$\pm$0.001 &  0.259$\pm$0.038 &  0.297$\pm$0.009 &  0.178$\pm$0.037 &  0.147$\pm$0.013 &  0.129$\pm$0.025 &  0.114$\pm$0.017 \\
&TERM & 0.177$\pm$0.034 &  0.137$\pm$0.025 &    0.4$\pm$0.066 &  0.415$\pm$0.006 &  0.303$\pm$0.074 &   0.33$\pm$0.029 &  0.172$\pm$0.036 &  0.172$\pm$0.076 &  0.164$\pm$0.044 &   0.18$\pm$0.005 \\

&FedMinMax (ours) &   0.261$\pm$0.007 &  0.184$\pm$0.007 &  0.321$\pm$0.021 &  \textbf{0.351$\pm$0.009} &  0.295$\pm$0.003 &  0.323$\pm$0.011 &  0.22$\pm$0.008 &  0.299$\pm$0.011 &  0.201$\pm$0.001 &  0.154$\pm$0.008  \\
\midrule

\multicolumn{2}{c}{Centalized Minmax Baseline} &  0.263$\pm$0.013 &  0.187$\pm$0.005 &  0.325$\pm$0.016 &  \textbf{0.352$\pm$0.003} &  0.293$\pm$0.007 &  0.334$\pm$0.017 &  0.216$\pm$0.009 &  0.305$\pm$0.009 &  0.205$\pm$0.002 &  0.154$\pm$0.005 \\

\bottomrule
\end{tabular}}
\label{analytic_cifar10} 
\end{table}

\begin{table}[H]
\centering
\caption{Final group weighting coefficients for AFL, Centalized Minmax Baseline, and FedMinmax across different federated learning scenarios on the CIFAR-10 dataset. The weights are rounded to the last three decimal places and the weighting coefficients for the worst group are in bold.
}
\resizebox{0.8\textwidth}{!}{
\begin{tabular}{llcccccccccc}
\toprule
 \textbf{Setting}&  \textbf{Method} &\textbf{Airplane} &  \textbf{Automobile} &      \textbf{Bird} &       \textbf{Cat} &      \textbf{Deer} &       \textbf{Dog} &      \textbf{Frog} &     \textbf{Horse} &      \textbf{Ship} & \textbf{Truck}\\
\midrule
 ESG &AFL &    0.100 &       0.100 &  0.100 &  0.101 &  0.099 &  0.102 &  0.099 &  0.100 &  0.100 &  0.100 \\
&FedMinMax (ours) &    0.075 &       0.031 &  0.152 &  \textbf{0.206}&  0.102 &  0.192 &  0.083 &  0.085 &  0.035 &  0.039 \\
\midrule
SSG &AFL &  0.088 &       0.031 &  0.140 &  \textbf{0.207} &  0.101 &  0.200 &  0.079 &  0.074 &  0.054 &  0.028 \\
&FedMinMax (ours) &0.071 &       0.030 &  0.147 &  \textbf{0.209} &  0.103 &  0.195 &  0.082 &  0.085 &  0.038 &  0.040 \\
\midrule
PSG & AFL &   0.091 &       0.066 &  0.119 &  0.128 &  0.118 &  0.103 &  0.102 &  0.097 &  0.081 &  0.097 \\
&FedMinMax (ours) &      0.078 &       0.045 &  0.143 & \textbf{0.207} &  0.108 &  0.203 &  0.080 &  0.078 &  0.033 &  0.024 \\
\midrule
\multicolumn{2}{c}{Centalized Minmax Baseline} & 0.082 & 0.017 &  0.139 &  \textbf{0.205} &  0.118 &  0.190 &  0.091 &  0.080 &  0.032 &  0.046\\

\bottomrule
\end{tabular}}
\label{tab:coeff_cifar} 
\end{table}

\bigskip

\noindent \textbf{Experiments on ACS Employment dataset (employment and race combination). } 
 In the \textit{Equal access to Sensitive Groups (ESG)} setting, we split the 6 groups across the clients equally. In the \textit{Partial access to Sensitive Groups (PSG)} setting,  20 clients own data from groups \textit{Unemployed White}, \textit{Employed Black}, \textit{Employed White}, and the remaining own data from \textit{Unemployed Other}, \textit{Unemployed Black}, and \textit{Employed Other}. Finally, in the \textit{Single access to Sensitive Groups (SSG)} setting, every client has access to only one sensitive class. In particular, data for \textit{Employed White} is owned by 10 clients and each of the remaining 5 groups is allocated to six clients.
 We report the risks on the test set in Table \ref{tab:analytic_acs_race_target_risks} and the group weighting coefficients produced from the training process are in Table \ref{tab:coeff_acs_employment_race_target}.

\begin{table}[H]
\centering
\caption{Test risks for FedAvg, AFL, q-FFL, TERM, and FedMinmax across different federated learning settings on ACS Employment dataset.}
\resizebox{\textwidth}{!}{
\begin{tabular}{llcccccc}
\toprule
 \textbf{Setting}&  \textbf{Method} &  \textbf{Unemployed White} &  \textbf{Employed White} &   \textbf{Employed Black} &  \textbf{Unemployed Other} &  \textbf{Unemployed Black} &  \textbf{Employed Other} 
 \\
\midrule
 ESG & AFL &  0.322$\pm$0.004 &   0.47$\pm$0.006 &   0.45$\pm$0.003 &  0.424$\pm$0.004 &  0.357$\pm$0.002 &  0.328$\pm$0.004  \\
&FedAvg &  0.312$\pm$0.003 &  0.486$\pm$0.005 &  0.459$\pm$0.002 &  0.435$\pm$0.004 &  0.351$\pm$0.002 &  0.317$\pm$0.003\\
&q-FedAvg &  0.335$\pm$0.005 &  0.451$\pm$0.007 &   0.44$\pm$0.004 &  0.411$\pm$0.005 &  0.365$\pm$0.003 &  0.341$\pm$0.006 \\
& TERM &   0.349$\pm$0.006 &  0.431$\pm$0.008 &  0.429$\pm$0.004 &  0.396$\pm$0.006 &  0.373$\pm$0.003 &  0.357$\pm$0.007 \\
&FedMinMax (ours) &    0.383$\pm$0.003 &  \textbf{0.374$\pm$0.005} &  0.381$\pm$0.001 &  0.366$\pm$0.008 &  0.374$\pm$0.001 &    0.36$\pm$0.01 \\
\midrule
SSG &AFL & 0.386$\pm$0.01 &   \textbf{0.374$\pm$0.004} &  0.384$\pm$0.007 &  0.365$\pm$0.009 &  0.377$\pm$0.009 &  0.362$\pm$0.007 \\
&FedAvg &  0.256$\pm$0.002 &  0.596$\pm$0.005 &  0.517$\pm$0.003 &  0.527$\pm$0.005 &  0.316$\pm$0.002 &  0.249$\pm$0.003  \\
&q-FedAvg & 0.261$\pm$0.003 &  0.582$\pm$0.007 &   0.51$\pm$0.005 &  0.513$\pm$0.007 &   0.32$\pm$0.002 &  0.258$\pm$0.004 \\
& TERM &  0.27$\pm$0.001 &  0.563$\pm$0.003 &  0.499$\pm$0.001 &  0.499$\pm$0.003 &  0.326$\pm$0.001 &  0.267$\pm$0.002 \\
&FedMinMax (ours) &    0.384$\pm$0.006 &   \textbf{0.373$\pm$0.004} &  0.383$\pm$0.003 &  0.365$\pm$0.005 &  0.375$\pm$0.007 &   0.36$\pm$0.007  \\

\midrule
PSG & AFL &0.287$\pm$0.004 &  0.529$\pm$0.008 &  0.481$\pm$0.005 &  0.469$\pm$0.005 &  0.337$\pm$0.003 &  0.289$\pm$0.003 \\
&FedAvg &   0.278$\pm$0.005 &  0.548$\pm$0.011 &  0.491$\pm$0.006 &  0.485$\pm$0.004 &  0.331$\pm$0.004 &  0.277$\pm$0.003  \\
&q-FedAvg &  0.296$\pm$0.002 &  0.513$\pm$0.003 &  0.472$\pm$0.002 &  0.457$\pm$0.003 &  0.343$\pm$0.001 &  0.298$\pm$0.003 \\
& TERM & 0.303$\pm$0.004 &    0.5$\pm$0.008 &  0.466$\pm$0.005 &  0.447$\pm$0.001 &  0.347$\pm$0.003 &  0.306$\pm$0.001  \\
&FedMinMax (ours) &  0.385$\pm$0.004 &  \textbf{0.375$\pm$0.005} &  0.384$\pm$0.006 &  0.364$\pm$0.001 &  0.376$\pm$0.003 &   0.36$\pm$0.002 \\
\midrule
\multicolumn{2}{c}{Centalized Minmax Baseline} & 0.381$\pm$0.006 &  \textbf{0.375$\pm$0.003} &  0.382$\pm$0.002 &  0.367$\pm$0.004 &  0.374$\pm$0.007 &  0.359$\pm$0.011 \\

\bottomrule
\end{tabular}}

\label{tab:analytic_acs_race_target_risks} 
\end{table}

\begin{table}[H]
\centering
\caption{Final group weighting coefficients for AFL, Centalized Minmax Baseline, and FedMinmax for the ACS Employment dataset. The weights are rounded to the last three decimal places.
}
\resizebox{0.99\textwidth}{!}{
\begin{tabular}{llccccccccc}
\toprule
 \textbf{Setting}&  \textbf{Method} &   \textbf{Unemployed White} &  \textbf{Employed White} &   \textbf{Employed Black} &  \textbf{Unemployed Other} &  \textbf{Unemployed Black} &  \textbf{Employed Other} \\
\midrule
 ESG &AFL & 0.419 & 0.351 & 0.038 &  0.078 &  0.062 & 0.052  \\
&FedMinMax (ours) &  0.461 & 0.356 & 0.041 &  0.044 &  0.070 & 0.029 \\
\midrule
SSG &AFL &  0.461 & 0.355 & 0.040 &  0.045 &  0.070 & 0.029  \\
&FedMinMax&  0.461 & 0.356 & 0.040 &  0.045 &  0.070 & 0.028\\
\midrule
PSG & AFL & 0.431 & 0.343 & 0.040 &  0.072 &  0.067 & 0.048 \\
&FedMinMax (ours) &  0.461 & 0.356 & 0.041 &  0.044 &  0.070 & 0.029  \\ 
\midrule
\multicolumn{2}{c}{Centalized Minmax Baseline} & 0.461 & 0.356 & 0.040 &  0.045 &  0.070 & 0.029  \\

\bottomrule
\end{tabular}}
\label{tab:coeff_acs_employment_race_target} 
\end{table}

\bigskip
\noindent \textbf{Experiments on ACS Employment dataset (race).} We also use the original 9 races of the ACS Employment dataset to run experiments on the three federated learning settings.
We refer to the available race groups using the following label tags: $\{$ \textit{White}: White alone, \textit{Black}: Black or African American alone, \textit{American Indian}: American Indian alone, \textit{Alaska Native}: Alaska Native alone, \textit{A.I. \&/or A.N. Tribes}: American Indian and Alaska Native tribes specified, or American Indian or Alaska Native, not specified and no other races, \textit{Asian}: Asian alone, \textit{N. Hawaiian \& other P.I.}: Native Hawaiian and Other Pacific Islander alone, \textit{Other}: Some Other Race alone, \textit{Multiple}: Two or More Races$\}$. In the \textit{Equal access to Sensitive Groups (ESG)} setting, we split the 9 groups across the clients. In the \textit{Partial access to Sensitive Groups (PSG)} setting,  20 clients own data from groups \textit{White}, \textit{Black /African American}, \textit{American Indian}, \textit{Alaska Native}, \textit{A.I. \&/or A.N. Tribes} and the remaining clients hold data from \textit{Asian}, \textit{N. Hawaiian \& other P.I.}, \textit{Other}, and \textit{Multiple}. Finally, in the \textit{Single access to Sensitive Groups (SSG)} setting, every client owns only one sensitive group and each group is distributed to only 4 clients, except \textit{White} race that is distributed to 8 clients.
 We report the risks on the test set in Table \ref{tab:analytic_acs_race_risks}.
 
\begin{table}[H]
\centering
\caption{Risks for FedAvg, AFL, q-FFL, TERM, and FedMinmax across different federated learning settings on ACS Employment dataset.
}
\resizebox{\textwidth}{!}{
\begin{tabular}{llccccccccc}
\toprule
 \textbf{Setting}&  \textbf{Method} &  \textbf{White} & \textbf{Black} &	\textbf{American Indian}	& \textbf{Alaska Native} &\textbf{	A.I. \&/or A.N. Tribes} &	\textbf{Asian} &	\textbf{N. Hawaiian \& other P.I.}&	\textbf{Other}&	\textbf{Multiple}
 \\
\midrule
 ESG & AFL &   0.47$\pm$0.003 &  0.477$\pm$0.002 &  0.499$\pm$0.002 &  0.438$\pm$0.009 &  0.555$\pm$0.001 &  0.487$\pm$0.001 &  0.526$\pm$0.003 &  0.468$\pm$0.006 &  0.363$\pm$0.001 \\
&FedAvg & 0.471$\pm$0.004 &  0.477$\pm$0.002 &  0.501$\pm$0.002 &  0.437$\pm$0.012 &  0.556$\pm$0.001 &  0.488$\pm$0.001 &  0.526$\pm$0.004 &  0.471$\pm$0.007 &  0.363$\pm$0.002 \\
&q-FedAvg & 0.47$\pm$0.001 &  0.476$\pm$0.001 &    0.499$\pm$0.0 &  0.436$\pm$0.005 &  0.554$\pm$0.001 &    0.487$\pm$0.0 &  0.525$\pm$0.001 &  0.468$\pm$0.002 &    0.363$\pm$0.0 \\
& TERM &  0.47$\pm$0.004 &  0.483$\pm$0.005 &  0.504$\pm$0.007 &  0.398$\pm$0.043 &  0.553$\pm$0.001 &  0.488$\pm$0.001 &  0.527$\pm$0.004 &  0.469$\pm$0.008 &  0.365$\pm$0.003 \\
&FedMinMax (ours) &   0.467$\pm$0.0 &  0.48$\pm$0.001 &  0.5$\pm$0.001 &  0.375$\pm$0.004 &  \textbf{0.545$\pm$0.0} &  0.487$\pm$0.001 &    0.522$\pm$0.0 &  0.464$\pm$0.001 &  0.363$\pm$0.0  \\
\midrule
SSG &AFL &  0.467$\pm$0.0 &   0.479$\pm$0.0 &  0.499$\pm$0.0 &  0.396$\pm$0.003 &  \textbf{0.547$\pm$0.001} &    0.488$\pm$0.0 &  0.523$\pm$0.0 &    0.465$\pm$0.0 &    0.362$\pm$0.0 \\
&FedAvg &   0.473$\pm$0.002 &  0.475$\pm$0.001 &  0.501$\pm$0.0 &  0.412$\pm$0.009 &  0.575$\pm$0.003 &  0.487$\pm$0.0 &  0.524$\pm$0.003 &  0.482$\pm$0.001 &  0.363$\pm$0.001 \\
&q-FedAvg & 0.472$\pm$0.001 &  0.475$\pm$0.001 &    0.5$\pm$0.0 &  0.418$\pm$0.005 &  0.571$\pm$0.001 &  0.487$\pm$0.0 &  0.525$\pm$0.001 &   0.48$\pm$0.001 &    0.364$\pm$0.0 \\
& TERM &  0.469$\pm$0.001 &    0.474$\pm$0.0 &  0.5$\pm$0.001 &  0.421$\pm$0.006 &  0.567$\pm$0.002 &  0.487$\pm$0.0 &  0.525$\pm$0.001 &   0.48$\pm$0.001 &    0.363$\pm$0.0 \\
&FedMinMax (ours) &   0.467$\pm$0.0 &  0.479$\pm$0.001 &  0.499$\pm$0.0 &  0.383$\pm$0.004 & \textbf{ 0.546$\pm$0.001} &  0.487$\pm$0.001 &    0.522$\pm$0.0 &  0.465$\pm$0.001 &    0.363$\pm$0.0 \\

\midrule
PSG & AFL & 0.468$\pm$0.0 &    0.475$\pm$0.0 &  0.503$\pm$0.002 &    0.424$\pm$0.0 &    0.563$\pm$0.0 &   0.49$\pm$0.001 &  0.529$\pm$0.002 &  0.481$\pm$0.002 &  0.365$\pm$0.001   \\
&FedAvg &   0.468$\pm$0.0 &  0.475$\pm$0.001 &  0.503$\pm$0.003 &  0.421$\pm$0.002 &  0.564$\pm$0.001 &  0.489$\pm$0.003 &  0.529$\pm$0.004 &  0.481$\pm$0.003 &  0.365$\pm$0.001  \\
&q-FedAvg & 0.468$\pm$0.0 &    0.475$\pm$0.0 &  0.503$\pm$0.001 &   0.43$\pm$0.011 &  0.561$\pm$0.001 &   0.49$\pm$0.001 &   0.53$\pm$0.002 &   0.48$\pm$0.002 &  0.365$\pm$0.001  \\
& TERM &  0.471$\pm$0.006 &  0.476$\pm$0.003 &  0.502$\pm$0.003 &  0.434$\pm$0.009 &  0.559$\pm$0.001 &  0.489$\pm$0.002 &  0.528$\pm$0.005 &  0.474$\pm$0.009 &  0.364$\pm$0.002  \\
&FedMinMax (ours) &  0.467$\pm$0.0 &   0.48$\pm$0.001 &    0.5$\pm$0.001 &  0.373$\pm$0.004 &  \textbf{0.546$\pm$0.001} &    0.486$\pm$0.0 &    0.522$\pm$0.0 &    0.465$\pm$0.0 &  0.363$\pm$0.001  \\
\midrule
\multicolumn{2}{c}{Centalized Minmax Baseline} &   0.467$\pm$0.0 &    0.48$\pm$0.0 &  0.5$\pm$0.001 &  0.372$\pm$0.002 & \textbf{ 0.545$\pm$0.0} &  0.486$\pm$0.001 &  0.522$\pm$0.001 &  0.465$\pm$0.001 &  0.364$\pm$0.0 \\

\bottomrule
\end{tabular}}
\label{tab:analytic_acs_race_risks} 
\end{table}

\section{Appendix: Complementary Algorithms}\label{compl_algorithms}
In the main text we refer to slightly different optimization objective and an algorithm that we use to compare the generalization efficiency of considering global demographics on some scenarios. Here we provide more information about this approach.
In particular we consider the following empirical objective:
\begin{equation}
\min_{\bm \theta \in \Theta} \max_{\bm\mu \in \Delta_{\ge \epsilon}^{|
\mathcal{A}||
\mathcal{K}| - 1}} \hat{r}_{a,k}(\bm \theta)  \coloneqq \sum_{a \in \mathcal{A}} \sum_{k \in \mathcal{K}}\mu_{a,k}\hat{r}_{a,k}(\bm \theta).
\label{local_empirical_fed_objective}
\end{equation}
Note that this optimization objective assumes that each demographic group that a client has access to is treated as a unique sensitive group even if the same group exists in several clients.
We extend our FedMinMax algorithm to solve the objective in Eq. \ref{local_empirical_fed_objective}. The adjusted algorithm is called LocalFedMinMax for which we share the pseudocode in Algorithm \ref{alg:localfedminmax} and the full table of risks for FashionMNIST and CIFAR-10 in Tables \ref{tab:local_global_fmnist} and \ref{tab:local_global_cifar10}, respectively.
LocalFedMinMax and FedMinMax behave similarly on the worst group on SSG regardless for different number of clients, while LocalFedMinMax has higher worst group risks for the remaining settings compared to FedMinMax.

\begin{algorithm}
    
    \caption{\textsc{Local Federated MiniMax (LocalFedMinMax) }}
    \label{alg:localfedminmax}
    
    {\bfseries Input:} $\mathcal{K}$: Set of clients, $T:$ total number of communication rounds,
     $\eta_{\bm \theta}$: model learning rate, $\eta_{\bm\mu}$: global adversary learning rate, $\mathcal{S}_{a,k}$: set of examples for group $a$ in client $k$, $\forall a \in \mathcal{A}$ and $\forall k \in \mathcal{K}$.
     
    \begin{algorithmic}[1]
    \STATE Server {\bfseries initializes} $\bm\mu^0\leftarrow \rho = \{\{|\mathcal{S}_{a,k}|/|\mathcal{S}| \}_{a \in \mathcal{A}}\}_{k\in \mathcal{K}}$ and $\bm \theta^0$ randomly.
    
    \FOR{$t=1$ {\bfseries to} $T$}

    \STATE Server {\bfseries computes} $\bm w^{t-1} \leftarrow \bm \mu^{t-1} / \rho$
    
    \STATE Server {\bfseries broadcasts} $\bm \theta^{t-1}$, $\bm w^{t-1}$ 

      \FOR{each client $k \in \mathcal{K}$ {\bfseries in parallel }}
        \STATE  $\bm \theta^{t}_k \leftarrow \bm \theta^{t-1} - \eta_\theta \nabla_\theta \hat{r}_k(\bm \theta^{t-1}, \bm w^{t-1})$%
     
        \STATE Client-$k$ {\bfseries obtains} and {\bfseries sends} $\{\hat{r}_{a,k}(\bm \theta^{t-1})\}_{a \in \mathcal{A}}$ and  $\bm \theta^{t}_k$ to server
        \ENDFOR 
        
    \STATE Server {\bfseries computes}:  $\bm \theta^{t} \leftarrow \sum\limits_{k \in \mathcal{K}} \frac{n_k}{n} \bm \theta^{t}_k$
     
     \STATE Server {\bfseries updates:}
     $\bm\mu^{t} \leftarrow \prod_{\Delta^{|\mathcal{K}|*|\mathcal{A}|-1}} \Big (\bm{{\mu}}^{t-1} +\eta_{\bm{\mu}} {\nabla}_{\bm{\mu}}\langle\ \bm \mu^{t-1},\hat{r}_{a,k}(\bm \theta^{t-1})\rangle \Big ) $ 
     \ENDFOR
    
    \end{algorithmic}
    
    {\bfseries Outputs}: $\frac{1}{T} \sum_{t =1}^T \bm \theta^t$
\end{algorithm}

\begin{table}[H]
\caption{Brier Score risks for FedMinMax and LocalFedMinMax on FashionMNIST across the different federated learning scenarios.}
\resizebox{\textwidth}{!}{
\begin{tabular}{llllllllllll}
\toprule
 \textbf{Setting} &  \textbf{Method} & \textbf{Airplane}  & \textbf{Automobile} & \textbf{Bird} & \textbf{Cat} & \textbf{Deer} & \textbf{Dog} & \textbf{Frog} & \textbf{Horse} & \textbf{Ship} & \textbf{Truck}\\

\midrule
 ESG &LocalFedMinMax &  0.298$\pm$0.054 &  0.173$\pm$0.021 &   \textbf{{0.316$\pm$0.092}} &  0.224$\pm$0.006 &  0.256$\pm$0.036 &  0.184$\pm$0.033 &   0.29$\pm$0.022 &  0.157$\pm$0.042 &  0.185$\pm$0.015 &  0.149$\pm$0.019 \\
(10 clients)&FedMinMax (ours) &   0.25$\pm$0.003 &  0.168$\pm$0.014 &  0.218$\pm$0.015 &  0.205$\pm$0.008 &  0.243$\pm$0.025 &  0.184$\pm$0.021 &  \textbf{0.31$\pm$0.005} &  0.159$\pm$0.016 &  0.174$\pm$0.017 &  0.143$\pm$0.005 \\
\midrule
SSG &LocalFedMinMax &  0.288$\pm$0.055 &  0.153$\pm$0.009 &  0.253$\pm$0.069 &   0.22$\pm$0.023 &  0.251$\pm$0.029 &  0.161$\pm$0.024 &  \textbf{0.309$\pm$0.013} &   0.15$\pm$0.021 &  0.166$\pm$0.007 &  0.135$\pm$0.004 \\

(10 clients)&FedMinMax (ours) &0.265$\pm$0.004 &  0.184$\pm$0.023 &  0.229$\pm$0.016 &  0.216$\pm$0.017 &  0.256$\pm$0.031 &  0.192$\pm$0.029 & \textbf{0.308$\pm$0.003} &  0.177$\pm$0.029 &  0.193$\pm$0.023 &  0.158$\pm$0.014 \\

\midrule
PSG & LocalFedMinMax &  \textbf{0.331$\pm$0.007} &  0.153$\pm$0.008 &   0.323$\pm$0.03 &  0.232$\pm$0.005 &     0.23$\pm$0.0 &  0.152$\pm$0.012 &  {0.307$\pm$0.012}&  0.131$\pm$0.005 &   0.167$\pm$0.01 &  0.134$\pm$0.003 \\
(10 clients)&FedMinMax (ours) &   0.266$\pm$0.002 &  0.187$\pm$0.021 &  0.278$\pm$0.029 &  0.217$\pm$0.015 &  0.201$\pm$0.044 &   0.192$\pm$0.04 &  \textbf{0.308$\pm$0.012} &  0.165$\pm$0.022 &  0.187$\pm$0.026 &  0.158$\pm$0.011 \\
\midrule
ESG &LocalFedMinMax &  0.284$\pm$0.008 &    0.03$\pm$0.012 &  \textbf{{{0.346$\pm$0.081}}} &  0.147$\pm$0.007 &  0.232$\pm$0.006 &     0.156$\pm$0.006 &  0.271$\pm$0.004 &  0.165$\pm$0.0 &    0.09$\pm$0.008 &   0.154$\pm$0.009 \\
(40 clients)&FedMinMax (ours) &  0.261$\pm$0.006 &  0.191$\pm$0.016 &  0.256$\pm$0.027 &  0.217$\pm$0.013 &  0.223$\pm$0.031 &  0.207$\pm$0.027 &   \textbf{{0.307$\pm$0.01}} &  0.172$\pm$0.016 &  0.193$\pm$0.021 &  0.156$\pm$0.011 \\
\midrule
SSG &LocalFedMinMax &  0.25$\pm$0.005 &  0.206$\pm$0.003 &  0.24$\pm$0.006 &  0.25$\pm$0.007 &  0.28$\pm$0.007 &  0.23$\pm$0.01 &  \textbf{{0.31$\pm$0.05}} &  0.105$\pm$0.006 &  0.18$\pm$0.008 &  0.182$\pm$0.001 \\

(40 clients)&FedMinMax (ours) &  0.269$\pm$0.012 &    0.2$\pm$0.026 &  0.238$\pm$0.017 &  0.231$\pm$0.013 &  0.252$\pm$0.034 &    0.2$\pm$0.024 & \textbf{{0.309$\pm$0.011}} &   0.177$\pm$0.03 &  0.205$\pm$0.032 &  0.169$\pm$0.013 \\
\midrule
PSG & LocalFedMinMax &  \textbf{{{0.331$\pm$0.021}} }&  0.039$\pm$0.006 &  0.281$\pm$0.001 &  0.178$\pm$0.006 &  0.191$\pm$0.051 &  0.065$\pm$0.05 &  0.275$\pm$0.006 &  0.068$\pm$0.1 &  0.041$\pm$0.09 &  0.12$\pm$0.2 \\
(40 clients)&FedMinMax (ours) &  0.263$\pm$0.013 &  0.177$\pm$0.026 &  0.228$\pm$0.011 &   0.21$\pm$0.019 &  0.238$\pm$0.025 &   0.182$\pm$0.03 &   \textbf{{0.31$\pm$0.008}} &   0.16$\pm$0.027 &  0.184$\pm$0.031 &  0.154$\pm$0.018 \\

\bottomrule
\end{tabular}}
\label{tab:local_global_fmnist} 
\end{table}

\begin{table}[H]
\caption{Brier score risks for LocalFedMinMax and FedMinmax on CIFAR-10 dataset across different federated learning scenarios.}
\resizebox{\textwidth}{!}{
\begin{tabular}{llllllllllll}
\toprule
 \textbf{Setting} &  \textbf{Method} & \textbf{Airplane}  & \textbf{Automobile} & \textbf{Bird} & \textbf{Cat} & \textbf{Deer} & \textbf{Dog} & \textbf{Frog} & \textbf{Horse} & \textbf{Ship} & \textbf{Truck}\\
 \midrule
ESG &LocalFedMinMax &   0.24$\pm$0.039 &  0.119$\pm$0.015 &  0.319$\pm$0.018 &  \textbf{0.358$\pm$0.008} &  0.278$\pm$0.024 &  0.276$\pm$0.022 &  0.264$\pm$0.001 &  0.197$\pm$0.029 &  0.213$\pm$0.111 &   0.14$\pm$0.053 \\
(10 clients)&FedMinMax (ours) & 0.279$\pm$0.028 &  0.243$\pm$0.089 &   0.32$\pm$0.019 &   \textbf{0.352$\pm$0.02} &   0.266$\pm$0.03 &   0.33$\pm$0.013 &   0.229$\pm$0.02 &  0.323$\pm$0.029 &  0.222$\pm$0.037 &  0.219$\pm$0.022 \\
\midrule
SSG &LocalFedMinMax &  0.263$\pm$0.012 &   0.236$\pm$0.04 &   0.227$\pm$0.09 &    \textbf{0.352$\pm$0.0} &   0.29$\pm$0.009 &  0.334$\pm$0.017 &  0.234$\pm$0.039 &   0.25$\pm$0.055 &  0.199$\pm$0.013 &  0.156$\pm$0.003 \\
(10 clients)& FedMinMax (ours) & 0.278$\pm$0.032 &  0.211$\pm$0.043 &  0.284$\pm$0.083 &  \textbf{0.351$\pm$0.0} &  0.287$\pm$0.004 &  0.328$\pm$0.008 &  0.213$\pm$0.014 &  0.267$\pm$0.065 &  0.204$\pm$0.002 &  0.157$\pm$0.009 \\

\midrule
PSG & LocalFedMinMax &  0.235$\pm$0.018 &  0.161$\pm$0.044 &  0.294$\pm$0.004 &  \textbf{0.353$\pm$0.042} &  0.249$\pm$0.073 &   0.331$\pm$0.03 &  0.226$\pm$0.025 &    0.236$\pm$0.0 &  0.189$\pm$0.096 &  0.223$\pm$0.093 \\
(10 clients)& FedMinMax (ours) &   0.236$\pm$0.024 &  0.185$\pm$0.006 &  0.334$\pm$0.004 &  \textbf{0.351$\pm$0.005} &  0.296$\pm$0.007 &  0.341$\pm$0.016 &  0.217$\pm$0.012 &  0.248$\pm$0.082 &   0.23$\pm$0.032 &  0.179$\pm$0.029 \\
 \midrule
ESG &LocalFedMinMax &    0.203$\pm$0.05 &  0.152$\pm$0.024 &  0.326$\pm$0.016 &  \textbf{0.381$\pm$0.004} &  0.304$\pm$0.069 &  0.335$\pm$0.045 &  0.195$\pm$0.065 &  0.171$\pm$0.027 &  0.167$\pm$0.086 &  0.182$\pm$0.063 \\
(40 clients)&FedMinMax (ours) &  0.257$\pm$0.003 &  0.189$\pm$0.009 &  0.324$\pm$0.015 &  \textbf{{0.351$\pm$0.002}} &  0.291$\pm$0.004 &   0.291$\pm$0.03 &  0.231$\pm$0.007 &  0.309$\pm$0.008 &  0.194$\pm$0.002 &  0.158$\pm$0.008  \\
\midrule
SSG &LocalFedMinMax &0.245$\pm$0.032 &  0.119$\pm$0.015 &  0.312$\pm$0.029 &  \textbf{0.352$\pm$0.007} &  0.298$\pm$0.004 &  0.307$\pm$0.066 &  0.235$\pm$0.039 &  0.226$\pm$0.013 &  0.275$\pm$0.025 &  0.233$\pm$0.079 \\
(40 clients)&FedMinMax (ours) &   0.258$\pm$0.01 &  0.187$\pm$0.005 &  0.332$\pm$0.005 &  \textbf{{0.351$\pm$0.002}} &  0.293$\pm$0.007 &  0.334$\pm$0.017 &  0.216$\pm$0.009 &  0.305$\pm$0.009 &  0.205$\pm$0.002 &  0.154$\pm$0.005  \\

\midrule
PSG & LocalFedMinMax &  0.236$\pm$0.027 &  0.14$\pm$0.038 &  0.32$\pm$0.025 &  \textbf{0.378$\pm$0.005} &  0.296$\pm$0.005 &  0.314$\pm$0.048 &  0.232$\pm$0.028 &  0.214$\pm$0.023 &  0.267$\pm$0.022 &  0.222$\pm$0.059 \\
(40 clients)& FedMinMax (ours) &   0.261$\pm$0.007 &  0.184$\pm$0.007 &  0.321$\pm$0.021 &  \textbf{{0.351$\pm$0.009}} &  0.295$\pm$0.003 &  0.323$\pm$0.011 &  0.22$\pm$0.008 &  0.299$\pm$0.011 &  0.201$\pm$0.001 &  0.154$\pm$0.008  \\

\bottomrule
\end{tabular}}
\label{tab:local_global_cifar10} 
\end{table}

\end{document}